\pdfoutput=1
\documentclass[11pt]{article}

\usepackage{authblk}

\usepackage{acl}

\usepackage{times}
\usepackage{latexsym}

\usepackage[T1]{fontenc}
\usepackage[utf8]{inputenc}

\usepackage{microtype}

\usepackage{amsmath}
\usepackage{amssymb}
\usepackage{graphicx}
\usepackage{makecell}
\usepackage{enumitem}
\usepackage{multirow}
\usepackage{booktabs}
\usepackage{etoolbox}
\usepackage{pgfplots}
\usepackage{pifont}% http://ctan.org/pkg/pifont

\usepackage{relsize}

\newcommand{\xmark}{\ding{53}}%
\pgfplotsset{compat=1.14, every non boxed x axis/.append style={x axis line style=-},
     every non boxed y axis/.append style={y axis line style=-}}
\usepgfplotslibrary{fillbetween}

\AtBeginEnvironment{quote}{\par\singlespacing\small}

\newenvironment{myquote}%
{\list{}{\leftmargin=0.3in}\item[]}%
{\endlist}

\usepackage[super]{nth}

\title{Claim-Dissector: An Interpretable Fact-Checking System with \\ Joint Re-ranking and Veracity Prediction}

\author[1, 2]{\textbf{Martin Fajcik}}
\author[1, 2]{\textbf{Petr Motlicek}}
\author[2]{\textbf{Pavel Smrz}}

\affil[1]{IDIAP Research Institute, Martigny, Switzerland}
\affil[2]{Brno University of Technology, Brno, Czech Republic}
\affil[ ]{\texttt{martin.fajcik@vut.cz}}
\begin{document}
\maketitle
\begin{abstract}
We present Claim-Dissector: a novel latent variable model for fact-checking and analysis, which given a claim and a set of retrieved evidences jointly learns to identify: (i) the relevant evidences to the given claim, (ii) the veracity of the claim. We propose to disentangle the per-evidence relevance probability and its contribution to the final veracity probability in an interpretable way --- the final veracity probability is proportional to a linear ensemble of per-evidence relevance probabilities. In this way, the individual contributions of evidences towards the final predicted probability can be identified. In per-evidence relevance probability, our model can further distinguish whether each relevant evidence is supporting (S) or refuting (R) the claim. This allows to quantify how much the S/R probability contributes to the final verdict or to detect disagreeing evidence.

Despite its interpretable nature, our system achieves results competitive with state-of-the-art on the FEVER dataset, as compared to typical two-stage system pipelines, while using significantly fewer parameters. It also sets new state-of-the-art on FAVIQ and RealFC datasets.
Furthermore, our analysis shows that our model can learn fine-grained relevance cues while using coarse-grained supervision, and we demonstrate it in 2 ways. (i) We show that our model can achieve competitive sentence recall while using only paragraph-level relevance supervision. (ii) Traversing towards the finest granularity of relevance, we show that our model is capable of identifying relevance at the token level. To do this, we present a new benchmark TLR-FEVER focusing on token-level interpretability ---  humans annotate tokens in relevant evidences they considered essential when making their judgment. Then we measure how similar are these annotations to the tokens our model is focusing on.\footnote{\url{https://github.com/KNOT-FIT-BUT/ClaimDissector}.}
\end{abstract}

\section{Introduction}
%General Motivation
Today's automated fact-checking systems are moving from predicting the claim's veracity by capturing the superficial cues of credibility, such as the way the claim is written, the statistics captured in the claim author's profile, or the stances of its respondents on social networks \cite{zubiaga2016analysing,derczynski-etal-2017-semeval,gorrell-etal-2019-semeval,fajcik-etal-2019-fit,li-etal-2019-eventai} towards evidence-grounded systems which, given a claim, identify relevant sources and then use these to predict the claim's veracity \cite{thorne-etal-2018-fever, jiang-etal-2020-hover,park-etal-2022-faviq}.
In practice, providing precise evidence turns out to be at least as important as predicting the veracity itself. Disproving a claim without linking it to factual evidence often fails to be persuasive and can even cause a ``backfire'' effect --- refreshing and strengthening the belief into an erroneous claim \cite{lewandowsky2012misinformation}\footnote{Further discussion in Appendix~\ref{app:lewandovsky}.}. 

%SOTA and limitations
For evidence-grounded fact-checking, most of the existing state-of-the-art systems \cite{jiang-etal-2021-exploring-listwise,stammbach-2021-evidence,khattab2021baleen} employ a 3-stage cascade approach; given a claim, they retrieve relevant documents, rerank relevant evidences (sentences, paragraphs or larger text blocks) within these documents, and predict the claim's veracity from the top-$k$ (usually~$k$=5) relevant evidences.

This comes with several drawbacks; firstly, \textit{the multiple steps of the system lead to error propagation}, i.e. the input to the last system might often be too noisy to contain any information. Some previous work focused on merging evidence reranking and veracity prediction into a single step
  \cite{ma-etal-2019-sentence,schlichtkrull-etal-2021-joint}.
  Secondly, in open-domain setting, \textit{number of relevant evidences can be significantly larger than~$k$}\footnote{e.g.,\textasciitilde3.7\,\% of FEVER's non-exhaustive annotations.}, especially when there is a lot of repeated evidence.
  Thirdly, in open-domain setting, \textit{sometimes there is both, supporting and refuting evidence}. The re-ranking systems often do not distinguish whether evidence is relevant because it supports or refutes the claim, and thus may select the evidence from one group based on the in-built biases.
%\todo{Mention not the greatest per-token evaluation protocol of previous work too.}

To further strengthen the persuasive effect of the evidences and understand the model's reasoning process, some of these systems provide cues of interpretability \cite{popat-etal-2018-declare,liu2020fine}. However, the interpretability in the mentioned work was often considered a useful trait, which was evaluated only qualitatively, as the labor-intensive human evaluation was out of the scope of their focus.

%Approach and Contribution.
To this extent, we propose Claim-Dissector (CD), a latent variable model which:
\begin{enumerate}
\item jointly ranks top-relevant, top-supporting and top-refuting evidences, and predicts veracity of the claim in an interpretable way, where the probability of the claim's veracity is estimated using the linear combination of per-evidence probabilities (Subsection \ref{ss:verifier}),
\item can provide fine-grained (sentence-level or token-level evidence), while using only coarse-grained supervision (on block-level or sentence-level respectively),
\item can be parametrized from a spectrum of language representation models (such as \mbox{RoBERTa} or \mbox{DeBERTaV3} \cite{liu2019roberta,he2021debertav3}).
\end{enumerate}
Finally, we collect a 4-way annotated dataset \texttt{TLR-FEVER} of per-token relevance annotations. This serves as a proxy for evaluating interpretability: we measure how similar are the cues provided by the model to the ones from humans. We believe future work can benefit from our quantitative evaluation approach while maintaining focus.

\section{Model Description}
\begin{figure}[t!]
    \centering
    \includegraphics[width=\linewidth]{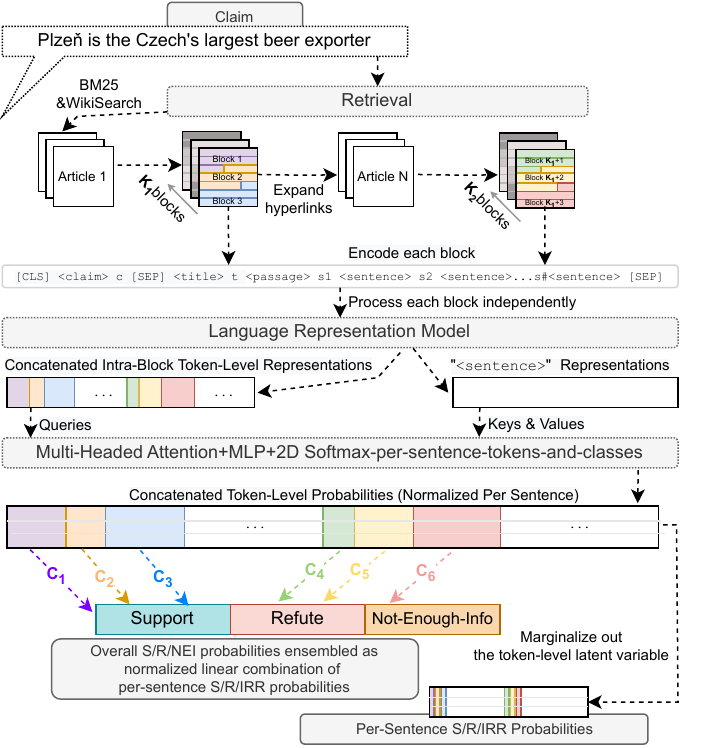}
    \caption{Diagram of Claim-Dissector's workflow. Abbreviations S, R, IRR, NEI stand for support, refute, irrelevant, not-enough-information. MLP and 2D softmax functions from the figure are defined in equation~\ref{eq:scorecomp} and equation~\ref{eq:word_label_probability} respectively.}
    \label{fig:main_schema}
\end{figure}
We present a 2-stage system composed of the \textit{retriever} and the \textit{verifier}. The documents are ranked via retriever. Each document is split into blocks. The blocks from top-ranking documents are passed to verifier and jointly judged. Our interpretable CD verifier is capable of re-ranking documents for any granularity of relevant evidence (e.g., document, block, sentence, token). Jointly, the same model predicts the claim's veracity. The overall schema of our approach is depicted in Figure~\ref{fig:main_schema}.

\subsection{Retriever}
\label{sec:retriever}
Given a claim~$c \in \mathcal{C}$ from the set of all possible claims~$\mathcal{C}$  and the corpus~$\mathcal{D}=\{d_1, d_2, ... , d_n\}$ composed of documents~$d_i$, the retriever produces a ranking using function~$\operatorname{rank}:\mathcal{C} \times \mathcal{D} \rightarrow \mathbb{R}$ that assigns a claim-dependent score to each document in the corpus. In this work, we focus on the verifier; therefore, we take the strong retriever from \citet{jiang-etal-2021-exploring-listwise}. This retriever interleaves documents ranked by BM25 \cite{robertson2009probabilistic} ($a_1,a_2,...a_n$) and Wikipedia API ($b_1,b_2,...b_m$) following \citet{hanselowski2018ukp} as ($a_1,b_1,a_2,b_2,...$), while skipping duplicate articles. Each document is then split into non-overlapping blocks of size~$L_x$, respecting sentence boundaries\footnote{Every block contains as many sentences as can fit into~$L_x$ tokens, considering verifier's tokenization.}. Our verifier then computes its veracity prediction from top-$K_1$ such blocks.
To keep up with similar approaches \cite{hanselowski2018ukp,stammbach-neumann-2019-team}%subramanian-lee-2020-hierarchical can also be cited here, if space is left
, we also experiment with expanding evidence with documents hyperlinked to the top retrieved articles. We rank these documents according to rank and sequential order in the document they were hyperlinked from. We then process these extra ranked documents the same way as retrieved documents, adding top-$K_2$ blocks to the verifier's input. As discussed more closely in \citet{stammbach-neumann-2019-team}, some relevant documents are impossible to retrieve using just claim itself, as their relevance is conditioned on other relevant documents. However, we stress that such approaches also mimic the way FEVER dataset was collected, and thus the improvements of such approach on ``naturally collected'' datasets might be negligible if any.

\subsection{Verifier}
\label{ss:verifier}
The verifier first processes each block independently using a language representation model (LRM) and then aggregates cross-block information via multi-head attention \cite{NIPS2017_transformers}, computing matrix~$\boldsymbol{M}$. This matrix is used to compute both, the probability of each evidence's relevance and the probability of the claim's veracity. Furthermore, the way the model is constructed allows learning a linear relationship between these probability spaces.

Formally given a claim~$c$  and~$K=K_1+K_2$ blocks,~$K$ input sequences~$x_i$ for each block~$i$ are constructed as
\begin{myquote}
{\small\texttt{[CLS] <claim> $c$ [SEP] <title>~$t$ <passage>~$s_1$ <sentence>~$s_2$ <sentence>...$s_\#$<sentence> [SEP]}},
\end{myquote}
where \texttt{[CLS]} and \texttt{[SEP]} are transformer special tokens used during the LRM pre-training \cite{devlin-etal-2019-bert}. 
Each block is paired with its article's title~$t$ and split into sentences~$s_1, s_2, ..., s_\#$. Symbols~$c, t, s_1, s_2, ..., s_\#$ thus each denote a sequence of tokens.
We further introduce new special tokens \texttt{<claim>, <title>, <passage>, <sentence>} to separate different input parts. 
Crucially, every sentence is appended with a \texttt{<sentence>} token. Their respective embeddings are trained from scratch. 
Each input~$x_i$ is then encoded via LRM~$\boldsymbol{E}_i=\operatorname{LRM}(x_i) \in \mathbb{R}^{L_B \times d}$, where~$L_B$ is an input sequence length, and~$d$ is LRM's hidden dimensionality. 
The representations of every block are then concatenated into~$\boldsymbol{E} = [\boldsymbol{E}_1; \boldsymbol{E}_2; ... ; \boldsymbol{E}_{K}] \in \mathbb{R}^{L \times d}$, where~$L$ is the number of all tokens in the input sequences from all retrieved blocks. 
Then we index-select all representations from~$\boldsymbol{E}$ corresponding to positions of sentence tokens in~$s_1, s_2, ..., s_\#$ into score matrix~$\boldsymbol{E_s} \in \mathbb{R}^{L_e \times d}$, where~$L_e$ corresponds to the number of all tokens in all input sentences (without special tokens).
Similarly, we index-select all representations at the same positions as the special \texttt{<sentence>} tokens at the input from~$\boldsymbol{E}$ into matrix~$\boldsymbol{S}\in \mathbb{R}^{L_S \times d}$, where~$L_S \ll L_e$ is the total number of sentences in all inputs~$x_i$. 
The matrix~$\boldsymbol{M} \in \mathbb{R}^{L_e \times 3}$ is then given as
\begin{equation}
\label{eq:scorecomp}
    \boldsymbol{M} = \operatorname{SLP}(\operatorname{MHAtt}(\boldsymbol{E}_s,\boldsymbol{S},\boldsymbol{S}))\boldsymbol{W}.
\end{equation}
The~$\operatorname{MHAtt}: \ldots \rightarrow\mathbb{R}^{L_e \times d}$ operator is a multi-head attention with queries~$\boldsymbol{E}_s$, and keys and values~$\boldsymbol{S}$. 
The~$\operatorname{SLP}$ operator is a single layer perceptron with layer norm, dropout, and GeLU activation (details in Appendix~\ref{sec:app_SLP}) and~$\boldsymbol{W} \in \mathbb{R}^{d \times 3}$ is a linear transformation, projecting resulting vectors to the desired number of classes (3 in case of FEVER).
% Footnote on the importance of not using a residual connection
% Based on 1 experiment observation. I will rather not report on this, unless I will have more reliable data.
%\footnote{Crucially, we do not use residual connection for cross-block layer from equation~\ref{eq:scorecomp}. In preliminary experiments, we found that using it yields barely any improvement over not using this layer at all. We hypothesize, that it was caused by model learning to ignore aggregated representations computed by the non-pretrained parameters, when given a chance.}. 

To compute the per-evidence probabilities we split the matrix~$\boldsymbol{M}$ according to tokens belonging to each evidence. For instance, for sentence-level evidence granularity we do split~$\boldsymbol{M}= [\boldsymbol{M}^{s_{1},1};\boldsymbol{M}^{s_{2},1};...;\boldsymbol{M}^{s_{\#},K}]$ along dimension~$L_e$ into submatrix representations corresponding to sentence~$s_1$ in block~$1$ up to last sentence~$s_\#$ in block~$K$. We then independently normalize each such matrix of~$i$-th evidence of~$j$-th block as\footnote{Note that the probability also depends on input sequences~$\{x_i\}_{i\in\{1,2,...,K\}}$, but we omit this dependency for brevity.}:
\begin{equation}
    \label{eq:word_label_probability}
    \operatorname{P}^{i,j}(w, y)=\frac{\operatorname{exp}{\boldsymbol{M}^{i,j}_{w,y}}}{\sum_{w'}\sum_{y'}\operatorname{exp}{\boldsymbol{M}^{i,j}_{w',y'}}}.
\end{equation}
Note that $w \in \{1,2,...,|s_{i,j}|\}$ is a token index in the (i,j)-th evidence and $y\in\{$S, R, NEI$\}$ is the class label.
Then we marginalize over latent variable~$w$ to obtain the marginal log-probability per evidence.
\begin{equation}
    \log\operatorname{P}^{i,j}(y)=\log\sum_{w'}\operatorname{P}^{i,j}(y, w')
\end{equation}
Then objective~$\mathcal{L}_R$ is computed for evidences annotated in label set~$\mathbb{A}=\{(y^*_1,(i_1,j_1)),...\}$ (usually $|\mathbb{A}|\ll L_S$) for a single claim\footnote{If example has NEI veracity in FEVER, $\mathcal{L}_R=0$.}.
\begin{equation}
\label{eq:loss0}
    \mathcal{L}_R=\frac{1}{|\mathbb{A}|}\sum_{y^*, (i,j) \in \mathbb{A}} \log\operatorname{P}^{i,j}(y^*)
\end{equation}
In training, $\mathbb{A}$ contains the same amount of relevant and irrelevant labels. For relevant, the log-probability~$\log\operatorname{P}^{i,j}({y}=y^*)$ is maximized, based the overall claim's veracity label $y^* \in \{S,R\}$. For irrelevant evidences, $y^* =$ IRR is maximized. As FEVER contains only annotation of relevant sentences, we follow the heuristic of \citet{jiang-etal-2021-exploring-listwise} and sample irrelevant sentences ranked between 50 and 200, in order to avoid maximizing the objective for false negatives. In test-time, we rank the evidence~$(i,j)$ according to its combined probability of supporting or refuting relevance~$score_{i,j}=\sum_{y\in\{S, R\}}\operatorname{P}^{i,j}(y)$.

Next, we compute the probability of the claim's veracity~${y} \in \{$S, R, NEI$\}$. First notice that scores in~$\boldsymbol{M}$ are logits (proof in Appendix \ref{app:logit_proof})
\begin{equation}
    \label{eq:logit_wordlabelprob}
    \boldsymbol{M}^{i,j}_{w, y} = \log(C^{i,j}\operatorname{P}^{i,j}(w, y)).
\end{equation}
Therefore, we use a learnable extra non-negative degree of freedom~$C^{i,j}$ to compute a linear ensemble\footnote{Assuming~${y}$=IRR=NEI.} producing the final probability
\begin{equation}
    \operatorname{P}(y) = \frac{\sum_{i,j,{w}}{C^{i,j}\operatorname{P}^{i,j}({w}, {y})}}
    {\sum_{{y}'}\sum_{i,j,{w}}{C^{i,j}\operatorname{P}^{i,j}({w}, {y}')}}.
\end{equation}

Lastly, we bias the model to focus only on some tokens in each evidence by enforcing an $\mathcal{L}_2$ penalty over the scores in~$\boldsymbol{M}$ by
\begin{equation}
    \mathcal{L}_2 = \frac{1}{L_e}||\boldsymbol{M}||_{F}^2,
\end{equation}
where~$||\cdot||_F$  denotes Frobenius norm. We show empirically that this objective leads to significantly better weakly-supervised token-level interpretability (Section~\ref{section:results}). Therefore the final per-sample loss with hyperparameters~$\lambda_R$,~$\lambda_2$ is
\begin{equation}
    \mathcal{L} = -\log\operatorname{P}({y}) - \lambda_R\mathcal{L}_R + \lambda_2\mathcal{L}_2.
\end{equation}
% This will likely need to get out of model description section
\subsection{Baseline}
\label{section:baseline}
Apart from previous work, we propose a baseline bridging the proposed system and the recent work of \citet{schlichtkrull-etal-2021-joint}. In order to apply this recent work for FEVER, we introduce a few necessary modifications\footnote{The necessity of these is explained in Appendix \ref{app:baselinemods}.}. We normalize all scores in~$\boldsymbol{M}$ to compute joint probability across all blocks
\begin{equation}
    %  \operatorname{P}^{i,j}({w}, {y})=\frac{\operatorname{exp}{\boldsymbol{M}^{i,j}_{{w},{y}}}}{\sum_{{w'}}\sum_{{y'}}\operatorname{exp}{\boldsymbol{M}^{i,j}_{{w'},{y'}}}}.
    \operatorname{P}({w}, {y}) = \frac{\operatorname{exp}{\boldsymbol{M}_{{w},{y}}}}{\sum_{{w'}}\sum_{{y'}}\operatorname{exp}{\boldsymbol{M}_{{w'},{y'}}}}.
\end{equation}
%In training time, we filter $\boldsymbol{M}$ to only contain representations of annotated evidences (both relevant and irrelevant), as there are likely many more unannotated relevant evidences at the input.
First, we marginalize out per-token probabilities in each evidence~${s}_{i,j}$.
\begin{equation}
    %  \operatorname{P}^{i,j}({w}, {y})=\frac{\operatorname{exp}{\boldsymbol{M}^{i,j}_{{w},{y}}}}{\sum_{{w'}}\sum_{{y'}}\operatorname{exp}{\boldsymbol{M}^{i,j}_{{w'},{y'}}}}.
    \operatorname{P}({s}_{i,j}, {y}) = \sum_{{w'} \in {s}_{i,j}}\operatorname{P}({w'}, {y}) 
\end{equation}
Using this sentence probability formulation, the objective is computed for every relevant evidence.
\begin{equation}
    \mathcal{L}_{b0} = \frac{1}{|\mathbb{A}_p|}\sum_{{s}_{i,j}, {y} \in \mathbb{A}_p} \log \operatorname{P}({s}_{i,j}, {y}) 
\end{equation}

Second, unlike \citet{schlichtkrull-etal-2021-joint}, we interpolate objective~$\mathcal{L}_{b0}$ with objective
\begin{equation}
    \mathcal{L}_{b1} = \log\operatorname{P}({y}) = \log\sum_{{s}_{i,j}} \operatorname{P}({s}_{i,j}, {y})
\end{equation}
by computing their mean. Like CD, we use $\mathcal{L}_{b1}$ objective to take advantage of examples from NEI class for which we have no annotation in $\mathbb{A}_p$ (and thus~$\mathcal{L}_{b0}$ is virtually set to~$0$).
Unlike CD, the annotations~$\mathbb{A}_p$  in~$\mathcal{L}_{b0}$  contain only relevant labels where~${y^*}\in\{S, R\}$\footnote{Maximizing NEI class for irrelevant sentences leads to inferior accuracy. This makes sense, since it creates ``tug-of-war'' dynamics between~$\mathcal{L}_{b0}$ and~$\mathcal{L}_{b1}$. The former objective tries to allocate mass of joint space in NEI class, since most documents are irrelevant, whereas the latter objective tries to allocate the mass in the dimension of labeled veracity class.}.

In order to not penalize non-annotated false negatives, we compute global distribution in~$\mathcal{L}_{b0}$ during training only from representations of tokens from labeled positive and negative sentences in~$\boldsymbol{M}$.
In test time, we rank evidences according to~$score_{i,j}=\sum_{{y}\in\{S, R\}} \operatorname{P}({s_{i,j}}, {y})$, and predict claim's veracity according to~$\operatorname{P}({y}) = \sum_{{s}_{i,j}} \operatorname{P}({s}_{i,j}, {y})$. We also considered different model parametrizations discussed in Appendix~\ref{sec:app_diffparam}.

\subsection{Transferring Supervision to Higher Language Granularity}
The proposed model can benefit from annotation on the coarser granularity of the language than tested. For example, evidence annotation can be done at the document, block, paragraph, or token level. In Section~\ref{section:results}, we show despite the fact that the model is trained on coarse granularity level, the model still shows moderate performance of relevance prediction when evaluated on finer granularity. We demonstrate this with two experiments.

First, \textit{the model is trained with sentence-level supervision and it is evaluated on a token-level annotation.} For this we leave model as it is --- reminding that prior over per-token probabilities enforced by the objective~$\mathcal{L}_2$ is crucial (Table \ref{tab:slr_to_tlr}). 

Secondly, \textit{we assume only block-level annotation is available in training and we evaluate on sentence-level annotation}. Here, we slightly alter the model, making it rely more on its sentence-level representations.  In Section~\ref{section:results}, we show this simple alteration significantly improves the performance at sentence-level. To compute block-level probability, the block is the evidence, therefore the evidence index can be dropped. The probability of the~$j$-th block~${b}_{j}$ is obtained by marginalizing out the per-token/per-sentence probabilities.

\begin{equation}
\begin{split}
    \operatorname{P}({b}_{j}, {y}) = \sum_{{s}_{i,j} \in {b}_{j}}\operatorname{P}({s}_{i,j}, {y}) = \\ \sum_{{s}_{i,j} \in {b}_{j}} \sum_{{w'} \in {s}_{i,j}}\operatorname{P}({w'}, {y}) 
\end{split}
\end{equation}
In practice, we found it helpful to replace the block-level probability~$\operatorname{P}({b}_{j}, {y})$ with its lower-bound~$\operatorname{P}({s}_{i,j}, {y})$ computed for 1 sentence sampled from the relevant sentence likelihood.
\begin{equation}
\label{eq:approx_block}
    \operatorname{P}({b}_{j}, {y}) \approx \operatorname{P}({s}_{i,j}, {y}); {s}_{i,j} \sim \operatorname{p}(\boldsymbol{s}_{i,j}, {y}\in \{S,R\})
\end{equation}
Intuitively, making a single sentence estimate (SSE) forces the model to invest the mass into a few sentence-level probabilities. This is similar to \mbox{HardEM}\footnote{In preliminary experiments, we also tried \mbox{HardEM}, but the results over multiple seeds were unstable.}. In~$\mathcal{L}_R$ we then maximize the probabilities of positive blocks computed as in equation \ref{eq:approx_block}, and negative sentences\footnote{Indices of irrelevant sentences are mined automatically (see Section~\ref{sec:retriever}), therefore this supervision comes ``for free''.} computed (and normalized) on sentence level as in equation \ref{eq:loss0}.

\subsubsection{Baseline for Token-level Rationales}
Similarly to \citet{shah2020automatic, schuster-etal-2021-get}, we train a masker --- a model which learns to replace the least amount of token embeddings at the Claim-Dissector's input with a single learned embedding in order to maximize the NEI class probability. We compare the unsupervised rationales given by the masker with the unsupervisedly learned rationales provided by the Claim-Dissector on-the-fly. Our masker follows the same architecture as Claim-Dissector. We provide an in-depth description of our masker model and its implementation in Appendix~\ref{app:masker}.
%% Normalization is done on lower-granularity evidence level, and then the best lower-granularity evidence per higher-level granularity evidence is selected "as correct".

\section{Related Work}
\begin{description}[style=unboxed,leftmargin=0em,listparindent=\parindent]
    \setlength\parskip{0em}
\item[Datasets.]
Previous work in supervised open-domain fact-checking often focused on large datasets with evidence available in Wikipedia such as FEVER \cite{thorne-etal-2018-fever}, FEVER-KILT \cite{petroni-etal-2021-kilt}, FAVIQ \cite{park-etal-2022-faviq},  HoVer \cite{jiang-etal-2020-hover} or TabFact \cite{2019TabFactA}. We follow this line of work and selected FEVER because of its sentence-level annotation, 3 levels of veracity (into S/R/NEI classes), and controlled way of construction --- verification should not require world knowledge, everything should be grounded on trusted, objective, and factual evidence from Wikipedia.

\item[Open-Domain Fact-Checking (ODFC)]
Unlike this work, most of the previous work includes 3-stage systems that retrieve evidence, rerank each document independently, and then make a veracity decision from top-$k$ documents \cite{thorne-etal-2018-fever,nie2019combining, zhong-etal-2020-reasoning}.

\citet{jiang-etal-2021-exploring-listwise} particularly distinguished the line of work which aggregates the final decision from independently computed per-sentence veracity probabilities \cite[\textit{inter alia}]{zhou-etal-2019-gear,soleimani2020bert,pradeep2021vera} and the line of work where the top-relevant sentences are judged together to compute the final veracity probability \cite[\textit{inter alia}]{stammbach-neumann-2019-team, pradeep-etal-2021-scientific}. \citet{jiang-etal-2021-exploring-listwise} compares similar system against these two assumptions, showing that joint judgment of relevant evidence is crucial when computing final veracity. We stress that our system falls into joint judgment category. Although relevance is computed per sentence, these probabilities along with linear combination coefficients are computed jointly, with the model conditioned on hundreds of input sentences.

To deal with multi-hop evidence (evidence which is impossible to mark as relevant without other evidence) \citet{subramanian-lee-2020-hierarchical,stammbach-2021-evidence} iteratively rerank evidence sentences to find minimal evidence set, which is passed to verifier. Our system jointly judges sentences within a block, while multi-head attention layer could propagate cross-block information. Our overall peformance results suggest that our system is about on-par with these iterative approaches, while requiring only single forward computation. However, further analysis shows our model underperforms on multi-hop evidence (more in Section~\ref{section:results}).

%Interpretability
\item[Interpretability]
\citet{popat-etal-2018-declare, liu2020fine} both introduced systems with an interpretable attention design and demonstrated their ability to highlight important words through a case study. In our work, we take a step further and propose a principled way to evaluate our system quantitatively. We note that \citet{schuster-etal-2021-get} proposed a very similar quantitative evaluation of token-level rationales, for data from the VitaminC dataset. The dataset, constructed from factual revisions on Wikipedia, assumed that the revised part of facts is the most salient part of the evidence. In contrast, we instruct annotators to manually annotate terms important to their judgment (Section \ref{sec:tlrfever}).  The VitaminC dataset is not accompanied by the evidence corpus, thus we deemed it as unsuitable for open-domain knowledge processing.

\citet{krishna2021proofver} proposed a system that parses evidence sentences into natural logic based inferences \cite{angeli-manning-2014-naturalli}. These provide deterministic proof of the claim's veracity. Authors verify the interpretability of the generated proofs by asking humans to predict veracity verdict from them. However, the model is evaluated only on FEVER dataset and its derivatives, which contain potential bias to their approach --- the claims in this dataset were created from fact through "mutations" according to natural logic itself.

\item[Joint Reranking and Veracity Prediction]
\citet{schlichtkrull-etal-2021-joint} proposed a system similar to our work for fact-checking over tables. The system computes a single joint probability space for all considered evidence tables. The dataset however contains only claims with true/false outcomes, typically supported by a single table. While our work started ahead of its publication, it can be seen as an extension of this system.

\end{description}
\section{Experimental Setup}
Unless said otherwise, we employ \mbox{DeBERTaV3} \cite{he2021debertav3} as LRM. In all experiments, we firstly pre-train model on MNLI \cite{mnli}. We use maximum block-length~$L_x=500$. Our recipe for implementation and model training is closely described in Appendix \ref{app:experimental_details}.
\subsection{Datasets}
\begin{description}[style=unboxed,leftmargin=0em,listparindent=\parindent]
    \setlength\parskip{0em}
\item[FEVER.]
We validate our approach on FEVER \cite{thorne-etal-2018-fever} and our newly collected dataset of token-level rationales.
FEVER is composed from claims constructed from Wikipedia. Each annotator was presented with an evidence sentence, and first sentence of articles from hyperlinked terms. In FEVER, examples in development set contain multi-way annotation of relevant sentences, i.e., each annotator selected set of sentences (evidence group) they considered relevant. To analyze performance of our components on samples that need multi-hop reasoning, we further created subsets of training/development set. FEVER$_{MH}$ contains only examples where all annotators agreed on that more than 1 sentence is required for verification, whereas FEVER$_{{MH}_{ART}}$ contains only examples, where all annotators agreed that \textit{sentences from different articles} are required for verification. As majority of examples of FEVER$_{MH}$ are from FEVER$_{{MH}_{ART}}$, we only evaluate on FEVER$_{MH}$. We include the subset statistics in Table~\ref{tab:dataset_statistics}.
\begin{table}[t]
    \centering
    \resizebox{\linewidth}{!}{\begin{tabular}{crll}
\hline
\multicolumn{1}{l}{} & \multicolumn{1}{c}{\textbf{FEVER}} & \multicolumn{1}{c}{\textbf{FEVER$_{MH}$}} & \multicolumn{1}{c}{\textbf{FEVER$_{{MH}_{ART}}$}} \\ \hline
\textbf{Train}       & 145,449                            & 12,958 (8.91\%)                           & 11,701 (8.04\%)                                 \\
\textbf{Dev}         & 19,998                             & 1204/19998 (6.02\%)                       & 1059/19998 (5.30\%)                             \\
\textbf{Test}         & 19,998                            & -                                         & -                        \\\hline
\end{tabular}}%
    \caption{FEVER dataset and its subsets.}
    \label{tab:dataset_statistics}
\end{table}
\item[TLR-FEVER]
\label{sec:tlrfever}
To validate token-level rationales, we collect our own dataset on random subset of validation set (only considering examples with gold sentence annotation). We collect 4-way annotated set of token-level rationales. The annotators were the colleagues with NLP background from our lab. We instruct every annotator via written guidelines, and then we had 1-on-1 meeting after annotating a few samples, verifying that contents of the guidelines were understood correctly. We let annotators annotate~$100$ samples, resolve reported errors manually, obtaining~$94$ samples with fine-grained token-level annotation. In guidelines, we simply instruct annotators to \textit{highlight minimal part of text they find important for supporting/refuting the claim. There should be such part in every golden sentence (unless annotation error happened).} The complete guidelines are available in Appendix~\ref{app:annotation_guidelines}.

To establish performance of average annotator, we compute the performance of each annotator compared to other annotators on the dataset, and then compute the average annotator performance. We refer to this as \textit{human baseline lower-bound}, as each annotator was compared to only 3 annotations, while the system is compared to 4 annotations. We measure performance via F$_1$ metric.
% Commented: (thus the performance of average annotator on 4 annotations would be equal or better)

\item[Other Datasets.]
We further validate our approach on FAVIQ-A \cite{park-etal-2022-faviq} and RealFC \cite{thorne-vlachos-2021-evidence} datasets (Appendix \ref{app:FAVIQ} and Appendix \ref{app:REALFC}), where it achieves state-of-the-art results and HoVer (Appendix \ref{app:hover_res}), where we demonstrate its limitations on multi-hop evidence.
\end{description}

\subsection{Evaluation}
\label{ss:evaluation}
\begin{description}[style=unboxed,leftmargin=0em,listparindent=\parindent]
    \setlength\parskip{0em}
% \item[Recall@Input (RaI).] We evaluate retrieval w.r.t. recall at model's input while considering different amount of K$_1$+K$_2$ blocks at the input, i.e. the score hit counts iff any annotated evidence group was matched in K$_1$+K$_2$ input blocks.
\item[Accuracy (A).] The proportion of correctly classified samples, disregarding the predicted evidence.
\item[Recall@5 (R@5).] The proportion of samples for which any annotated evidence group is matched within top-$5$ ranked sentences.
\item[FEVER-Score (FS).] The proportion of samples for which (i) any annotated evidence group is matched within top-5 ranked sentences, and (ii) the correct class is predicted.
\item[F$_1$ Score] measures unigram overlap between predicted tokens and reference tokens, disregarding articles. Having multiple references, the maximum F$_1$ between prediction and any reference is considered per-sample. Our implementation closely follows \citet{rajpurkar-etal-2016-squad}. 

In practice, both CD and masker model infer continuous scores capturing relevance for every token\footnote{We consider mask-class logits as scores for masker.}. When evaluating F$_1$, we select only tokens with scores greater than threshold~$\tau$. We tune the optimal threshold~$\tau$ w.r.t. F$_1$ on TLR-FEVER.
\end{description}

\section{Results}
\label{section:results}
We report results of base-sized models based on 3-checkpoint average. We train only a single large model. We further evaluate retrieval in Appendix \ref{app:retrieval_perofmrnace} as it is a non-essential part of our contribution.
\begin{description}[style=unboxed,leftmargin=0em,listparindent=\parindent]
    \setlength\parskip{0em}
    
%%
% Performance
%%
\begin{table}[t]
    \centering
    \scalebox{0.60}{\begin{tabular}{clccccc}
\cmidrule{1-7}
& \textbf{System}                & \textbf{FS} & \textbf{A} & \textbf{R@5} & \textbf{HA} & \textbf{\#$\theta$} \\ \cmidrule{1-7}
\multirow{18}{*}{\rotatebox[origin=c]{90}{Development Set}} 
& TwoWingOS {\small\cite{yin-roth-2018-twowingos}}               & 54.3        & 75.9       & 53.8         & \xmark        & ?                                \\
& HAN {\small\cite{ma-etal-2019-sentence}}                       & 57.1        & 72.0       & 53.6         & \checkmark    & ?                                \\
& UNC {\small\cite{nie2019combining}}                            & 66.5        & 69.7       & 86.8         & \checkmark& 408M                             \\
& HESM {\small\cite{subramanian-lee-2020-hierarchical}}          & 73.4        & 75.8       & 90.5         & \checkmark        & 39M                              \\
& KGAT{[}OR{]} {\small\cite{liu2020fine}}                        & 76.1        & 78.3       & 94.4         & \xmark   & 465M                             \\
& DREAM {\small\cite{zhong-etal-2020-reasoning}}                 & -           & 79.2       & 90.5         & \checkmark$^?$        & 487M                             \\
& T5 {\small\cite{jiang-etal-2021-exploring-listwise}}           & 77.8        & \textbf{81.3}       & 90.5         & \xmark        & 5.7B                             \\
& LF+D$_{XL}$ {\small\cite{stammbach-2021-evidence}}             & -           & -          & 90.8         & \xmark        & 1.2B                             \\
& LF$_{2-iter}$+D$_{XL}$ {\small\cite{stammbach-2021-evidence}}  & -           & -          & \textbf{93.6}& \checkmark        & 1.2B                                      \\
& ProofVer-MV {\small\cite{krishna2021proofver}}                 & 78.2        & 80.2       & -            & \checkmark        & 515M                             \\ 
& ProofVer-SB {\small\cite{krishna2021proofver}}                 & \textbf{79.1}        & 80.7       & \textbf{93.6}            & \checkmark        & 765M                             \\ \cmidrule{2-7} 
& Baseline$_{joint}$                                             & 75.2        & 79.8       & 90.0         & \xmark   & 187M                             \\
& Claim-Dissector\small{$_{RoBERTa}$}                            & 74.6        & 78.6       & 90.4         & \xmark   & 127M                             \\
& Claim-Dissector\small{$_{RoBERTaL}$}                           & 75.1        & 79.1       & 90.6            & \xmark& 360M                            \\
& Claim-Dissector\small{$_{RoBERTaL}$  \textbackslash{}w HE}     & 76.1        & 79.4       & 91.7         & \checkmark& 360M                            \\
& Claim-Dissector                                                & 76.2        & 79.5       & 91.5         & \xmark   & 187M                             \\
& Claim-Dissector \small{\textbackslash{}w HE}                   & 76.9        & 79.8       & 93.0         & \checkmark& 187M                            \\
& Claim-Dissector\small{$_{L}$}                                  & 76.9        & 80.4       & 91.8         & \xmark   & 439M                             \\

& Claim-Dissector\small{$_{L}$ \textbackslash{}w HE}    & \underline{78.0}& \underline{80.8}& \underline{93.3}& \checkmark     & 439M                    \\
& Claim-Dissector{\small$_{L}$ \textbackslash{}w HE {[}OR{]}}    & 78.9        & 81.2       & 94.7         & \checkmark        & 439M                    \\ \cmidrule{1-7}
\multirow{12}{*}{\rotatebox[origin=c]{90}{Test Set}} 
& KGAT {\small\cite{liu2020fine}}                                & 70.4        & 74.1       & -            & \xmark  & 465M                             \\
& DREAM {\small\cite{zhong-etal-2020-reasoning}}                 & 70.6        & 76.9       & -            & \checkmark$^?$        & 487M                             \\
& HESM {\small\cite{subramanian-lee-2020-hierarchical}}          & 71.5        & 74.6       & -            & \checkmark        & 58M                              \\
& ProofVer-MV {\small\cite{krishna2021proofver}}                 & 74.4        & 79.3       & -            & \checkmark        & 515M                             \\
& T5 {\small\cite{jiang-etal-2021-exploring-listwise}}           & 75.9        & 79.4       & -            & \xmark        & 5.7B                             \\
& LF$_{2-iter}$+D$_{XL}$ {\small\cite{stammbach-2021-evidence}}  & 76.8        & 79.2       & -            & \checkmark        & 1.2B                             \\
& ProofVer-SB {\small\cite{krishna2021proofver}}                 & \textbf{76.8}        & \textbf{79.5}       & -            & \checkmark        & 765M                             \\\cmidrule{2-7}
& Claim-Dissector\small{$_{RoBERTaL}$}                           & 73.1        & 76.4       & -            & \xmark   & 360M                             \\
& Claim-Dissector\small{$_{RoBERTaL}$ \textbackslash{}w HE}      & 74.3        & 77.8       & -            & \checkmark & 360M                           \\
& Claim-Dissector\small{$_{L}$}                                  & 74.7        & 78.5       & -            & \xmark   & 439M                             \\
& Claim-Dissector\small{$_{L}$ \textbackslash{}w HE}             & \underline{76.5}& \underline{79.3}& -   & \checkmark & 439M                           \\\cmidrule{1-7}

\end{tabular}
}%
    \caption{Performance on dev and test splits of FEVER. \#$\theta$ denotes number of parameters in the model. Model names suffixed with [OR](as Oracle) inject missing golden evidence into its input. Models using any kind of hyperlink-augmentation (HA) are marked. Our models with hyperlink expansion are suffixed with (\textbackslash{}w HE).  \textbf{Overall best} and \underline{our best} result are in bold and underlined respectively (disregarding oracle results).}
    \label{fig:main_results}
\end{table}

\item[Performance.] We compare the performance of our system with previous work in Table~\ref{fig:main_results}. Results marked with $?$ were unknown/uncertain, and unconfirmed by authors. We note that, apart from HAN \cite{ma-etal-2019-sentence}, all previous systems were considering two separate systems for reranking and veracity prediction. Next, we note that only ProofVer system uses additional data. It leverages rewritten-claim data for fact-correction paired with original FEVER claims \cite{thorne-vlachos-2021-evidence}. 

We observe that (i) even our base-sized RoBERTa-based CD model outperforms base-sized HESM on dev data, and its large version outperforms large-sized KGAT, DREAM and HESM on test data, (ii) our base sized DebertaV3-based CD model matches large-based DREAM and even KGAT with oracle inputs on dev set, (iii) model version with hyperlink expansion (suffixed {\small\textbackslash{}w HE}) further improves the overall performance, (iv) using larger model improves mostly its accuracy, (v) Claim-Dissector{\small $_{L}$ \textbackslash{}w HE} achieves better FEVER score than T5-based approach (with two 3B models) and better accuracy than LongFormer+DebertaXL, but it is not pareto optimal to these previous SOTA, (vi) our model is outmatched by recent ProofVer-SB, though it is more efficient as ProofVer-SB requires two rounds of reranking and autoregressive decoding. We still consider this a strong feat, as our system was focusing on modeling reranking and veracity prediction jointly in an interpretable way.
Finally, we inject blocks with missing golden evidence into inputs of Claim-Dissector{\small$_{L}$ \textbackslash{}w HE} at random positions and measure oracle performance. We observe that items missed by retrieval are still beneficial to the performance.
%%
% Ablations
%%
\begin{table}[t]
    \centering
    \scalebox{0.66}{% \begin{tabular}{lccc|ccc}

\begin{tabular}{lccc|ccc}
                                              & \multicolumn{3}{c}{\textbf{FEVER}}                             & \multicolumn{3}{c}{\textbf{FEVER$_{MH}$}}      \\ \Xhline{4\arrayrulewidth}
\textbf{System}                               & \textbf{FS} & \textbf{A} & \textbf{R@5} & \textbf{FS} & \textbf{A} & \textbf{R@5} \\ \hline
CD$_{LARGE}$ \textbackslash{}w HE {[}OR{]}    & 78.9        & 81.2       & 94.8      & 50.3       & 81.9      & 58.9           \\
CD$_{LARGE}$ \textbackslash{}w HE             & 78.0        & 80.8       & 93.4      & 44.7       & 81.2      & 53.1           \\
CD \textbackslash{}w HE                       & 76.9        & 79.8       & 93.2      & 41.3       & 80.8      & 49.9           \\
CD \textbackslash{}w HE \textbackslash{}wo MH & 76.5        & 79.5       & 93.0      & 41.7       & 80.8      & 50.2           \\
Baseline                                      & 75.2        & 79.8       & 90.0      & 28.9       & 80.9      & 34.7           \\
CD                                            & 76.2        & 79.6       & 91.7      & 30.0       & 79.2      & 36.4           \\
CD \textbackslash{}wo $\mathcal{L}_2$         & 76.0        & 79.7       & 91.6      & 30.4       & 79.5      & 36.2           \\
CD \textbackslash{}wo VC                      & -           & -          & 91.9      & -          & -         & 37.8           \\
CD \textbackslash{}wo RC                      & -           & 79.9       & -         & -          & 81.5      & -              \\ \hline
\end{tabular}}%
    \caption{Ablation Study. Minor differences to Table \ref{fig:main_results} are caused by different early-stopping (Appendix \ref{app:experimental_details}).}
    \label{fig:ablations}
\end{table}
\item[Ablations.]
We ablate components of Claim-Dissector (CD) in Table~\ref{fig:ablations}. Firstly, we resort to single-task training. We drop veracity classification (VC) objective $\log\operatorname{P}({y})$  or relevance classification (RC) objective $\mathcal{L}_R$ from the loss. We observe an overall trend --- single-task model performs slightly better to multi-task model. The advantages of multi-task model, however, lie in its efficiency and ability to provide explanations between per-evidence relevances and final conclusion.
Next, we observe that dropping the~$\mathcal{L}_2$ objective doesn't affect the performance significantly. Further, we study the effect of hyperlink expansion (HE) and the effect of multi-head (MH) attention layer. As expected, hyperlink expansion dramatically increases performance on FEVER$_{MH}$. The multi-head attention also brings marginal improvements to results on FEVER. However, contrary to our expectations, there is no effect of the MH layer on FEVER$_{MH}$, the improvements happened in non-multihop part. Additional experiments with CD on HoVer dataset (Appendix \ref{app:hover_res}) confirm this, CD does not work well on examples with cross-article multihop evidence. Improving cross-article reasoning was not our aim, and we leave the investigation to our future research.

%%
% Token-level Interpretability
%%
\begin{table}[t]
    \centering
    \scalebox{0.7}{\begin{tabular}{lc}
\hline
\textbf{System}                       & \textbf{F1} \\ \hline
Select All Tokens                     & 52          \\
Select Claim Overlaps                 & 63          \\
Masker                                & 71          \\
Claim-Dissector \textbackslash{}wo $\mathcal{L}_2$ & 60          \\
Claim-Dissector                       & 77          \\
Human Performance LB                  & 85          \\\hline
\end{tabular}}%
    \caption{Token-level relevance on TLR-FEVER.}
    \label{tab:slr_to_tlr}
\end{table}
% Hidden for Review

\item[Transferring sentence-level supervision to token-level performance.]
We evaluated the performance of token-level rationales on our dataset in Table~\ref{tab:slr_to_tlr}. We considered two baselines. The first was to select all tokens in golden evidences (Select All Tokens). The second was to select only tokens that overlap with claim tokens (Select Claim Overlaps). We found that our model with weakly-supervised objective produces token-level rationales significantly better\footnote{See Appendix \ref{app:stat_testing_f1} for our F1 significance testing protocol.} than the masker --- a separate model trained explicitly to identify tokens important to the model's prediction. We hypothesize that a possible reason for this improvement could be that ad-hoc explainibility methods, that perturb the model's inputs (e.g., masker method or integrated gradients \cite{sundararajan2017axiomatic}) \textit{expose the model to out-of-distribution samples, thus degrading its performance and confidence}. Contrastively, this isn't happening by design with in-built interpretability methods like ours. Furthermore, the results also demonstrate the importance of~$\mathcal{L}_2$ objective. However, human performance is still considerably beyond the performance of our approach. 
%One such sample is shown in Appendix \ref{app:rationale_example}.

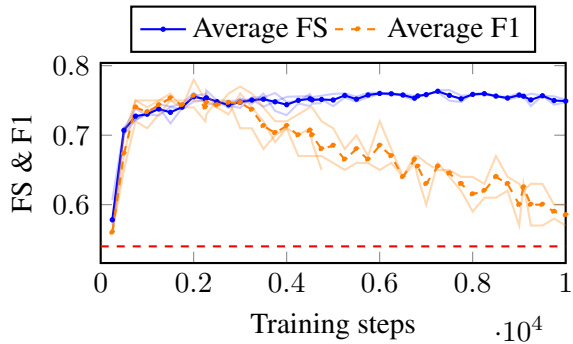
\begin{figure}
    \centering
    \begin{tikzpicture}
    \begin{axis}[
            %smooth,
            thick,
            legend columns=2,
            legend style={
                at={(0.5,1.05)}, 
                legend cell align={left},  
                anchor=south
            },
            xmin=0, xmax=10000,
            xlabel={Training steps},
            ylabel=FS \& F1,
            width=1.0\columnwidth,
            height=0.55\columnwidth,
            % xtick=data,
            % xmajorgrids,
            % ymajorgrids,
        ]%Steps,FS1,FS2,FS3,Average FS,F11,F12,F13,Average F1
        \addplot+[solid, mark=None, blue,  opacity=0.2,forget plot] table [x=Steps, y=FS1, col sep=comma] {data/data_training.csv};
        \addplot+[solid, mark=None, blue,  opacity=0.2,forget plot] table [x=Steps, y=FS2, col sep=comma] {data/data_training.csv};
        \addplot+[solid, mark=None, blue,  opacity=0.2,forget plot] table [x=Steps, y=FS3, col sep=comma] {data/data_training.csv};
        \addplot+[solid, mark=*, blue,mark options={scale=0.3}]               table [x=Steps, y=Average FS, col sep=comma] {data/data_training.csv};
        \addlegendentry{Average FS}

        \addplot+[solid, mark=None, orange,  opacity=0.3,forget plot] table [x=Steps, y=F11, col sep=comma] {data/data_training.csv};
        \addplot+[solid, mark=None, orange,  opacity=0.3,forget plot] table [x=Steps, y=F12, col sep=comma] {data/data_training.csv};
        \addplot+[solid,mark=None, orange,  opacity=0.3,forget plot] table [x=Steps, y=F13, col sep=comma] {data/data_training.csv};
        \addplot+[dashed,mark=*, orange,mark options={scale=0.35}]               table [x=Steps, y=Average F1, col sep=comma] {data/data_training.csv};
        \addlegendentry{Average F1}

        \addplot[dashed,thick, samples=50, smooth,domain=0:6,red] coordinates {(0,0.54)(10000,0.54)};
        % \addlegendentry{sentences + query}
        % \addlegendentry{sentences only (unst)}
        % \addlegendentry{sentences + query (unst)}
    \end{axis}%
\end{tikzpicture}%
    \caption{Average FEVER-Score (FS) and F1 performance on dev sets during training. Red dashed horizontal line marks the F1 performance when selecting all tokens (Select All Tokens) from Table \ref{tab:slr_to_tlr}. Opaque lines show the performance of individual checkpoints.}
    \label{fig:training_progress}
\end{figure}
In Figure \ref{fig:training_progress}, we analyze how the performance on FEVER-Score and F1 changes over the course of training on FEVER and TLR-FEVER sets. We find our scores reach the top performance and then detoriate. It is thus necessary do the early stopping on both, the performance metric and the interpretability metric. Furthermore, our experiments shown that tuning of $\lambda_2$ is crucial, i.e., $\lambda_2=2e-3$ tuned for DeBERTa-base, achieves no interpretability for the large version (where we use $5e-4$)\footnote{The sensitivity to $\lambda_2$ is analyzed in Appendix \ref{app:l2_sensitivity}.}. 

\begin{figure}
    \centering
    \includegraphics[width=\linewidth]{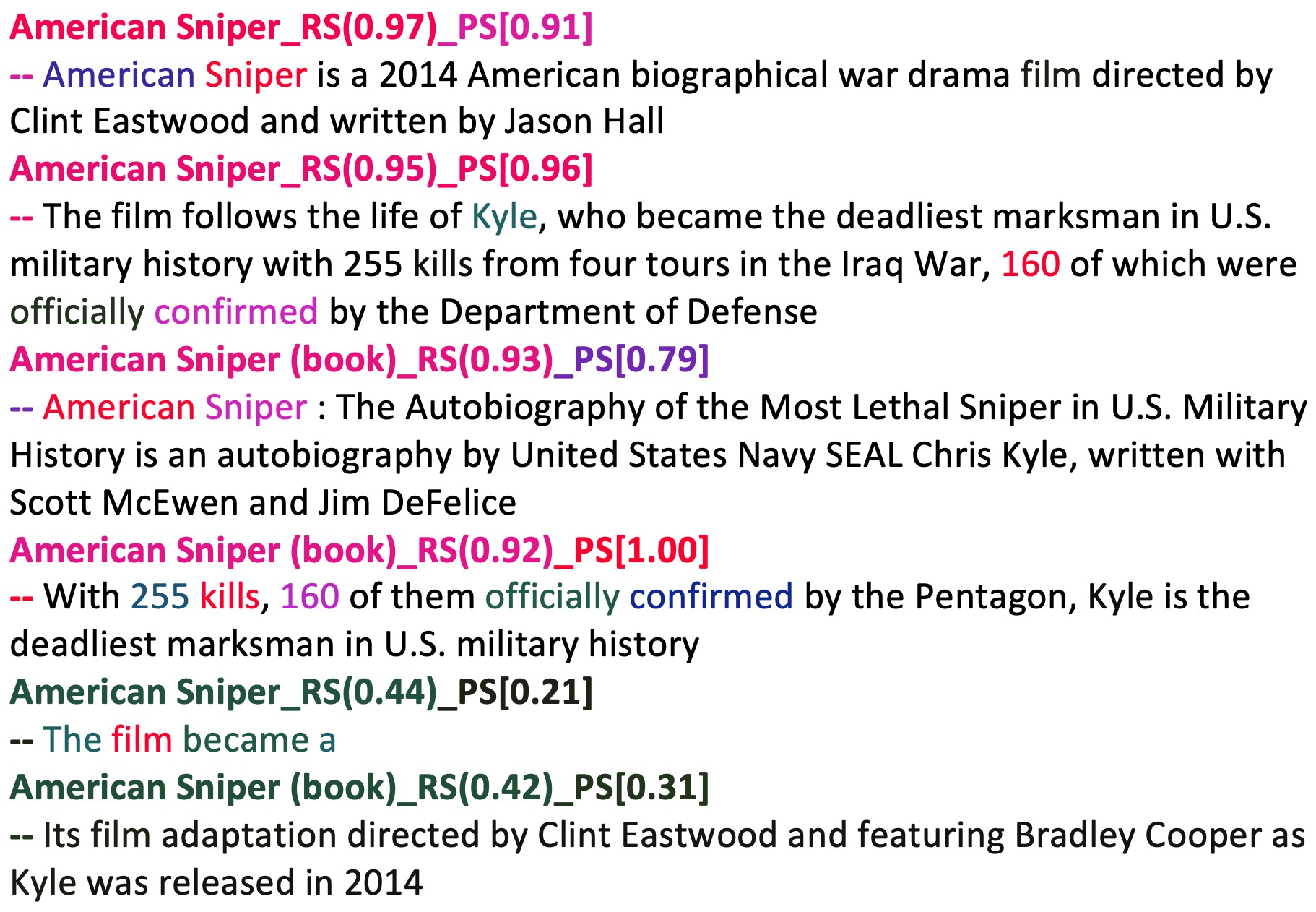}
    \caption{Qualitative sample of interpretable refuting evidence from Claim-Dissector for claim ``\textit{American Sniper (book) is about a sniper with 170 officially confirmed kills.}''.}
    \label{fig:intexample}
    \vspace{-3.5mm}
\end{figure}

Interpretable refuting example\footnote{We visualized token-level rationales on 100 random dev set examples at \url{shorturl.at/beTY2}.} is available in Figure~\ref{fig:intexample}. Example shows top-6 refuting sentences ranked by their refuting relevance probability (relevance score) (RS) $\operatorname{P}^{i,j}({y}=R)$. Each sentence is prefixed with its Wikipedia article title, RS and prediction score (PS). The prediction score is the corresponding non-negative linear coefficient $C^{i,j}$ max-normalized between 0 and 1 based on maximum $C^{i,j}$ for this sample. The token-level relevance, sentence relevance, and sentence prediction score are highlighted on a red-to-black scale (low relevance is black, high relevance is red). Interestingly, the prediction score is highest for sentences containing crucial refuting evidence --- the number of confirmed kills.

\item[Are prediction scores useful?]
In Figure \ref{fig:intexample}, a maximum prediction score is assigned to a sentence, that has a lower relevance score than the most relevant document. However, we argue that the sentence with the highest prediction score contains the most relevant information. Hence we formulate the hypothesis that \textit{top prediction score better ranks relevance towards final judgment, whereas top relevance score only reflects the model's confidence of the sentence being somehow relevant}. First, we note that scores are highly correlated, but not the same (average Kendall-$\tau$ 0.84). Next, we turn to the A/B testing (details in Appendix \ref{app:psuseful}), where we select 100 random samples such that: (i) each was correctly predicted, (ii) has a verdict \textit{supported} or \textit{refuted}. From these, we select (a) a sentence with the highest prediction score and (b) a sentence with equal or better relevance probability than (a); if there is no such sentence, we don't include it. We employ 5 annotators to say their preference for sentence (a), sentence (b), or neither (c).

We find that (i) 80\,\% of annotators unanimously agreed on \textit{not preferring} (b), (ii) 3 or more annotators in 73\,\% of cases preferred (a) over (b,c) and finally (iii) the worst single annotator preference for (a) over (a+b) cases was 86\,\%, demonstrating that human preferences strongly support declared hypothesis.

%\todo{Mention Relevance Score vs Prediciton Score}

% Technical Detail
% First we extract top rankings according relevance scores, with cutoff at RS=0.7. We assign each item in the list with PS and compute Kedall T between two rankins. We do this for each non-NEI dataset sample, and compute average Kendall-T. 

% MAIN TEXT
% We compute average kendall tau between rankings produced by 
% 0.8437366009778665

%\todo{Add discussion and data that validate prediction scores.}

\begin{table}[t]
    \centering
    \scalebox{0.8}{\begin{tabular}{lccc}
\hline
\textbf{Model}          & \textbf{FS} & \textbf{A} & \textbf{R@5} \\ \hline
Full Supervision        & 76.2        & 79.5       & 91.5         \\
Block Supervision       & 65.5        & 76.9       & 77.8         \\
Block Supervision + SSE & 69.7        & 78.1       & 83.0         \\ \hline
\end{tabular}}%
    \caption{Sentence-level performance on FEVER dev set under different kinds of supervision.}
    \label{fig:blr_to_slr}
\end{table}
\item[Transferring block-level supervision to sentence-level performance.]
The performance of our model on the sentence-level evidences is evaluated in Table~\ref{fig:blr_to_slr}. We notice that even our vanilla Claim-Dissector trained with block supervision reaches competitive recall@5 on sentence-level. However, adding SSE from equation \ref{eq:approx_block} leads to further improvements both in recall, but also in accuracy. We expected the recall will be improved, because model now focuses on assigning high probability mass only to some sentences within block, since high-entropy of the per-sentence distribution would be penalized in loss. However, we have not foreseen the damaging effect on accuracy, which block-level supervision causes. Interestingly, the accuracy without any evidence supervision reported in last row of Table~\ref{fig:ablations} was increased.

\end{description}

\section{Conclusion}
In this work, we proposed Claim-Dissector, an interpretable probabilistic model for fact-checking and fact-analysis. Our model jointly predicts the supporting/refuting evidence and the claim's veracity. It achieves state-of-the-art results, while providing
three layers of interpretability. Firstly, it identifies salient tokens important for the final prediction. Secondly, it allows disentangling ranking of relevant evidences into ranking of supporting evidence and ranking of refuting evidence. This allows detecting bipolar evidence without being exposed to such bipolar evidence sets during training (Appendix \ref{app:bipolarevi}). Thirdly, it combines the per-token relevance probabilities via linear combination into final veracity assessment. Therefore it can be identified to what extent the relevance of each token/sentence/block/document contributes to final assessment. Conveniently, this allows to differentiate between the concept of evidence relevance and its contribution to the final assessment. Our work was however limited in experiments with these coefficients, and we would like to analyze what they can learn, and how to inject features, such as satire assessment or source trustworthiness, through these coefficients in our future work.

Finally, it was shown that a hierarchical structure of our model allows making predictions on even finer language granularity, than the granularity the model was trained on. We believe the technique we proposed is transferable beyond fact-checking.

% These chapters do not count towards the page limit!
\section*{Ethical Considerations}
We built the system with the intended use for fact-checking support, providing rationales at various level to user for ease of understanding. These rationales include supporting and refuting evidences in the corpus. We see potential misuse of system might lie in spreading of fake news and propaganda by e.g., automatic detection of sources, which support or refute certain claims from the narrative. This could be followed by subsequent glorification/discreditation of statements in these sources. This could influence the human population, but also poison retrieval databases of similar fact-checking systems, influencing their decisions. Future work in this direction, such as \citet{Du_Bosselut_Manning_2022}, needs to study disinformation attacks and how to prevent them.

\section*{Limitations}
% Problems with block-level bias.
By manual analysis, we found that claim-dissector suffers from overconfidence in blocks with at least 1 relevant evidence. Then it seeks to select more relevant evidences inside, even when they are not. We believe this is connected to how irrelevant negatives are mined in FEVER --- they originate only from blocks without relevant evidences.

% Limitations with Commonsense Knowledge
On real data, the system often struggles to recognize what facts are refuting, and what are irrelevant (especially when applied out-of-domain). We demonstrate this in a case study on downstream application, where we replaced retrieval on Wikipedia with news-media in test-time. We tried to verify the claim "\textit{Weapons are being smuggled into Estonia}". Our system discovered article with facts about "\textit{Weapons being smuggled into Somalia}", and used it as a main refuting evidence to predict \texttt{REFUTE} veracity.

% Limitatations with Source credibility and trustworthiness
Lastly, CD is trained with evidence from Wikipedia, and do not considers other factors important for relevance assessment in practice, such as credibility of source, its reliability, or its narrative. This is the area of active research, as human fact-checkers also need to deal with lies \cite{uscinski2013epistemology}.

\section*{Acknowledgements}
We would like to thank to Santosh Kesiraju, Martin Dočekal, Karel Ondřej, Bolaji Yusuf, Michael Schlichtkrull, Juan Pablo Zulag-Gomez, and Apoorv Vyas for reviewing this work or providing helpful discussions. We also thank to Jungsoo Park and Sewon Min for their prompt responses and help with analysis of FAVIQ-A results.

This work was supported by the Technology Agency of the Czech Republic, project MASAPI (grant no. FW03010656), by CRiTERIA, an EU project, funded under the Horizon 2020 programme, grant agreement no. 101021866, and by the Ministry of Education, Youth and Sports of the Czech Republic through the e-INFRA CZ (ID:90140).

\bibliography{custom}

\begin{thebibliography}{58}
\expandafter\ifx\csname natexlab\endcsname\relax\def\natexlab#1{#1}\fi

\bibitem[{Angeli and Manning(2014)}]{angeli-manning-2014-naturalli}
Gabor Angeli and Christopher~D. Manning. 2014.
\newblock \href {https://doi.org/10.3115/v1/D14-1059} {{N}atural{LI}: Natural
  logic inference for common sense reasoning}.
\newblock In \emph{Proceedings of the 2014 Conference on Empirical Methods in
  Natural Language Processing ({EMNLP})}, pages 534--545, Doha, Qatar.
  Association for Computational Linguistics.

\bibitem[{Asai et~al.(2022)Asai, Gardner, and
  Hajishirzi}]{asai-etal-2022-evidentiality}
Akari Asai, Matt Gardner, and Hannaneh Hajishirzi. 2022.
\newblock \href {https://doi.org/10.18653/v1/2022.naacl-main.162}
  {Evidentiality-guided generation for knowledge-intensive {NLP} tasks}.
\newblock In \emph{Proceedings of the 2022 Conference of the North American
  Chapter of the Association for Computational Linguistics: Human Language
  Technologies}, pages 2226--2243, Seattle, United States. Association for
  Computational Linguistics.

\bibitem[{Ba et~al.(2016)Ba, Kiros, and Hinton}]{ba2016layer}
Jimmy~Lei Ba, Jamie~Ryan Kiros, and Geoffrey~E Hinton. 2016.
\newblock Layer normalization.
\newblock \emph{arXiv preprint arXiv:1607.06450}.

\bibitem[{Bishop(2006)}]{bishop:2006:PRML}
Christopher~M. Bishop. 2006.
\newblock \emph{Pattern Recognition and Machine Learning}.
\newblock Springer.

\bibitem[{Chen et~al.(2020)Chen, Wang, Chen, Zhang, Wang, Li, Zhou, and
  Wang}]{2019TabFactA}
Wenhu Chen, Hongmin Wang, Jianshu Chen, Yunkai Zhang, Hong Wang, Shiyang Li,
  Xiyou Zhou, and William~Yang Wang. 2020.
\newblock Tabfact : A large-scale dataset for table-based fact verification.
\newblock In \emph{International Conference on Learning Representations
  (ICLR)}, Addis Ababa, Ethiopia.

\bibitem[{Derczynski et~al.(2017)Derczynski, Bontcheva, Liakata, Procter, Wong
  Sak~Hoi, and Zubiaga}]{derczynski-etal-2017-semeval}
Leon Derczynski, Kalina Bontcheva, Maria Liakata, Rob Procter, Geraldine Wong
  Sak~Hoi, and Arkaitz Zubiaga. 2017.
\newblock \href {https://doi.org/10.18653/v1/S17-2006} {{S}em{E}val-2017 task
  8: {R}umour{E}val: Determining rumour veracity and support for rumours}.
\newblock In \emph{Proceedings of the 11th International Workshop on Semantic
  Evaluation ({S}em{E}val-2017)}, pages 69--76, Vancouver, Canada. Association
  for Computational Linguistics.

\bibitem[{Devlin et~al.(2019)Devlin, Chang, Lee, and
  Toutanova}]{devlin-etal-2019-bert}
Jacob Devlin, Ming-Wei Chang, Kenton Lee, and Kristina Toutanova. 2019.
\newblock \href {https://doi.org/10.18653/v1/N19-1423} {{BERT}: Pre-training of
  deep bidirectional transformers for language understanding}.
\newblock In \emph{Proceedings of the 2019 Conference of the North {A}merican
  Chapter of the Association for Computational Linguistics: Human Language
  Technologies, Volume 1 (Long and Short Papers)}, pages 4171--4186,
  Minneapolis, Minnesota. Association for Computational Linguistics.

\bibitem[{Du et~al.(2022)Du, Bosselut, and Manning}]{Du_Bosselut_Manning_2022}
Yibing Du, Antoine Bosselut, and Christopher~D. Manning. 2022.
\newblock \href {https://doi.org/10.1609/aaai.v36i10.21302} {Synthetic
  disinformation attacks on automated fact verification systems}.
\newblock \emph{Proceedings of the AAAI Conference on Artificial Intelligence},
  36(10):10581--10589.

\bibitem[{Fajcik et~al.(2019)Fajcik, Smrz, and Burget}]{fajcik-etal-2019-fit}
Martin Fajcik, Pavel Smrz, and Lukas Burget. 2019.
\newblock \href {https://doi.org/10.18653/v1/S19-2192} {{BUT}-{FIT} at
  {S}em{E}val-2019 task 7: Determining the rumour stance with pre-trained deep
  bidirectional transformers}.
\newblock In \emph{Proceedings of the 13th International Workshop on Semantic
  Evaluation}, pages 1097--1104, Minneapolis, Minnesota, USA. Association for
  Computational Linguistics.

\bibitem[{Gorrell et~al.(2019)Gorrell, Kochkina, Liakata, Aker, Zubiaga,
  Bontcheva, and Derczynski}]{gorrell-etal-2019-semeval}
Genevieve Gorrell, Elena Kochkina, Maria Liakata, Ahmet Aker, Arkaitz Zubiaga,
  Kalina Bontcheva, and Leon Derczynski. 2019.
\newblock \href {https://doi.org/10.18653/v1/S19-2147} {{S}em{E}val-2019 task
  7: {R}umour{E}val, determining rumour veracity and support for rumours}.
\newblock In \emph{Proceedings of the 13th International Workshop on Semantic
  Evaluation}, pages 845--854, Minneapolis, Minnesota, USA. Association for
  Computational Linguistics.

\bibitem[{Goutte and Gaussier(2005)}]{goutte2005probabilistic}
Cyril Goutte and Eric Gaussier. 2005.
\newblock A probabilistic interpretation of precision, recall and f-score, with
  implication for evaluation.
\newblock In \emph{European conference on information retrieval}, pages
  345--359. Springer.

\bibitem[{Hanselowski et~al.(2018)Hanselowski, Zhang, Li, Sorokin, Schiller,
  Schulz, and Gurevych}]{hanselowski2018ukp}
Andreas Hanselowski, Hao Zhang, Zile Li, Daniil Sorokin, Benjamin Schiller,
  Claudia Schulz, and Iryna Gurevych. 2018.
\newblock Ukp-athene: Multi-sentence textual entailment for claim verification.
\newblock In \emph{Proceedings of the First Workshop on Fact Extraction and
  VERification (FEVER)}, pages 103--108.

\bibitem[{He et~al.(2021)He, Gao, and Chen}]{he2021debertav3}
Pengcheng He, Jianfeng Gao, and Weizhu Chen. 2021.
\newblock \href {http://arxiv.org/abs/2111.09543} {Debertav3: Improving deberta
  using electra-style pre-training with gradient-disentangled embedding
  sharing}.

\bibitem[{Hendrycks and Gimpel(2016)}]{hendrycks2016gaussian}
Dan Hendrycks and Kevin Gimpel. 2016.
\newblock Gaussian error linear units (gelus).
\newblock \emph{arXiv preprint arXiv:1606.08415}.

\bibitem[{Izacard and Grave(2021)}]{izacard-grave-2021-leveraging}
Gautier Izacard and Edouard Grave. 2021.
\newblock \href {https://doi.org/10.18653/v1/2021.eacl-main.74} {Leveraging
  passage retrieval with generative models for open domain question answering}.
\newblock In \emph{Proceedings of the 16th Conference of the European Chapter
  of the Association for Computational Linguistics: Main Volume}, pages
  874--880, Online. Association for Computational Linguistics.

\bibitem[{Jang et~al.(2017)Jang, Gu, and Poole}]{jang2017categorical}
Eric Jang, Shixiang Gu, and Ben Poole. 2017.
\newblock Categorical reparametrization with gumble-softmax.
\newblock In \emph{International Conference on Learning Representations (ICLR
  2017)}. OpenReview. net.

\bibitem[{Jiang et~al.(2021)Jiang, Pradeep, and
  Lin}]{jiang-etal-2021-exploring-listwise}
Kelvin Jiang, Ronak Pradeep, and Jimmy Lin. 2021.
\newblock \href {https://doi.org/10.18653/v1/2021.acl-short.51} {Exploring
  listwise evidence reasoning with t5 for fact verification}.
\newblock In \emph{Proceedings of the 59th Annual Meeting of the Association
  for Computational Linguistics and the 11th International Joint Conference on
  Natural Language Processing (Volume 2: Short Papers)}, pages 402--410,
  Online. Association for Computational Linguistics.

\bibitem[{Jiang et~al.(2020)Jiang, Bordia, Zhong, Dognin, Singh, and
  Bansal}]{jiang-etal-2020-hover}
Yichen Jiang, Shikha Bordia, Zheng Zhong, Charles Dognin, Maneesh Singh, and
  Mohit Bansal. 2020.
\newblock \href {https://doi.org/10.18653/v1/2020.findings-emnlp.309}
  {{H}o{V}er: A dataset for many-hop fact extraction and claim verification}.
\newblock In \emph{Findings of the Association for Computational Linguistics:
  EMNLP 2020}, pages 3441--3460, Online. Association for Computational
  Linguistics.

\bibitem[{Karpukhin et~al.(2020)Karpukhin, Oguz, Min, Lewis, Wu, Edunov, Chen,
  and Yih}]{karpukhin-etal-2020-dense}
Vladimir Karpukhin, Barlas Oguz, Sewon Min, Patrick Lewis, Ledell Wu, Sergey
  Edunov, Danqi Chen, and Wen-tau Yih. 2020.
\newblock \href {https://doi.org/10.18653/v1/2020.emnlp-main.550} {Dense
  passage retrieval for open-domain question answering}.
\newblock In \emph{Proceedings of the 2020 Conference on Empirical Methods in
  Natural Language Processing (EMNLP)}, pages 6769--6781, Online. Association
  for Computational Linguistics.

\bibitem[{Khattab et~al.(2021)Khattab, Potts, and Zaharia}]{khattab2021baleen}
Omar Khattab, Christopher Potts, and Matei Zaharia. 2021.
\newblock Baleen: Robust multi-hop reasoning at scale via condensed retrieval.
\newblock \emph{Advances in Neural Information Processing Systems}, 34.

\bibitem[{Krishna et~al.(2022)Krishna, Riedel, and
  Vlachos}]{krishna2021proofver}
Amrith Krishna, Sebastian Riedel, and Andreas Vlachos. 2022.
\newblock Proofver: Natural logic theorem proving for fact verification.
\newblock \emph{Transactions of the Association for Computational Linguistics},
  10:1013--1030.

\bibitem[{Lewandowsky et~al.(2012)Lewandowsky, Ecker, Seifert, Schwarz, and
  Cook}]{lewandowsky2012misinformation}
Stephan Lewandowsky, Ullrich~KH Ecker, Colleen~M Seifert, Norbert Schwarz, and
  John Cook. 2012.
\newblock Misinformation and its correction: Continued influence and successful
  debiasing.
\newblock \emph{Psychological science in the public interest}, 13(3):106--131.

\bibitem[{Li et~al.(2019)Li, Zhang, and Si}]{li-etal-2019-eventai}
Quanzhi Li, Qiong Zhang, and Luo Si. 2019.
\newblock \href {https://doi.org/10.18653/v1/S19-2148} {event{AI} at
  {S}em{E}val-2019 task 7: Rumor detection on social media by exploiting
  content, user credibility and propagation information}.
\newblock In \emph{Proceedings of the 13th International Workshop on Semantic
  Evaluation}, pages 855--859, Minneapolis, Minnesota, USA. Association for
  Computational Linguistics.

\bibitem[{Liu et~al.(2019)Liu, Ott, Goyal, Du, Joshi, Chen, Levy, Lewis,
  Zettlemoyer, and Stoyanov}]{liu2019roberta}
Yinhan Liu, Myle Ott, Naman Goyal, Jingfei Du, Mandar Joshi, Danqi Chen, Omer
  Levy, Mike Lewis, Luke Zettlemoyer, and Veselin Stoyanov. 2019.
\newblock Roberta: A robustly optimized bert pretraining approach.
\newblock \emph{arXiv preprint arXiv:1907.11692}.

\bibitem[{Liu et~al.(2020)Liu, Xiong, Sun, and Liu}]{liu2020fine}
Zhenghao Liu, Chenyan Xiong, Maosong Sun, and Zhiyuan Liu. 2020.
\newblock Fine-grained fact verification with kernel graph attention network.
\newblock In \emph{Proceedings of the 58th Annual Meeting of the Association
  for Computational Linguistics}, pages 7342--7351.

\bibitem[{Loshchilov and Hutter(2017)}]{loshchilov2017decoupled}
Ilya Loshchilov and Frank Hutter. 2017.
\newblock Decoupled weight decay regularization.
\newblock \emph{arXiv preprint arXiv:1711.05101}.

\bibitem[{Ma et~al.(2019)Ma, Gao, Joty, and Wong}]{ma-etal-2019-sentence}
Jing Ma, Wei Gao, Shafiq Joty, and Kam-Fai Wong. 2019.
\newblock \href {https://doi.org/10.18653/v1/P19-1244} {Sentence-level evidence
  embedding for claim verification with hierarchical attention networks}.
\newblock In \emph{Proceedings of the 57th Annual Meeting of the Association
  for Computational Linguistics}, pages 2561--2571, Florence, Italy.
  Association for Computational Linguistics.

\bibitem[{Min et~al.(2020)Min, Michael, Hajishirzi, and
  Zettlemoyer}]{min-etal-2020-ambigqa}
Sewon Min, Julian Michael, Hannaneh Hajishirzi, and Luke Zettlemoyer. 2020.
\newblock \href {https://doi.org/10.18653/v1/2020.emnlp-main.466} {{A}mbig{QA}:
  Answering ambiguous open-domain questions}.
\newblock In \emph{Proceedings of the 2020 Conference on Empirical Methods in
  Natural Language Processing (EMNLP)}, pages 5783--5797, Online. Association
  for Computational Linguistics.

\bibitem[{Nie et~al.(2019)Nie, Chen, and Bansal}]{nie2019combining}
Yixin Nie, Haonan Chen, and Mohit Bansal. 2019.
\newblock Combining fact extraction and verification with neural semantic
  matching networks.
\newblock In \emph{Proceedings of the Thirty-Third AAAI Conference on
  Artificial Intelligence and Thirty-First Innovative Applications of
  Artificial Intelligence Conference and Ninth AAAI Symposium on Educational
  Advances in Artificial Intelligence}, pages 6859--6866.

\bibitem[{Park et~al.(2022)Park, Min, Kang, Zettlemoyer, and
  Hajishirzi}]{park-etal-2022-faviq}
Jungsoo Park, Sewon Min, Jaewoo Kang, Luke Zettlemoyer, and Hannaneh
  Hajishirzi. 2022.
\newblock \href {https://doi.org/10.18653/v1/2022.acl-long.354} {{F}a{VIQ}:
  {FA}ct verification from information-seeking questions}.
\newblock In \emph{Proceedings of the 60th Annual Meeting of the Association
  for Computational Linguistics (Volume 1: Long Papers)}, pages 5154--5166,
  Dublin, Ireland. Association for Computational Linguistics.

\bibitem[{Pascanu et~al.(2013)Pascanu, Mikolov, and
  Bengio}]{pascanu2013difficulty}
Razvan Pascanu, Tomas Mikolov, and Yoshua Bengio. 2013.
\newblock On the difficulty of training recurrent neural networks.
\newblock In \emph{International conference on machine learning}, pages
  1310--1318. PMLR.

\bibitem[{Petroni et~al.(2021)Petroni, Piktus, Fan, Lewis, Yazdani, De~Cao,
  Thorne, Jernite, Karpukhin, Maillard, Plachouras, Rockt{\"a}schel, and
  Riedel}]{petroni-etal-2021-kilt}
Fabio Petroni, Aleksandra Piktus, Angela Fan, Patrick Lewis, Majid Yazdani,
  Nicola De~Cao, James Thorne, Yacine Jernite, Vladimir Karpukhin, Jean
  Maillard, Vassilis Plachouras, Tim Rockt{\"a}schel, and Sebastian Riedel.
  2021.
\newblock \href {https://doi.org/10.18653/v1/2021.naacl-main.200} {{KILT}: a
  benchmark for knowledge intensive language tasks}.
\newblock In \emph{Proceedings of the 2021 Conference of the North American
  Chapter of the Association for Computational Linguistics: Human Language
  Technologies}, pages 2523--2544, Online. Association for Computational
  Linguistics.

\bibitem[{Popat et~al.(2018)Popat, Mukherjee, Yates, and
  Weikum}]{popat-etal-2018-declare}
Kashyap Popat, Subhabrata Mukherjee, Andrew Yates, and Gerhard Weikum. 2018.
\newblock \href {https://doi.org/10.18653/v1/D18-1003} {{D}e{C}lar{E}:
  Debunking fake news and false claims using evidence-aware deep learning}.
\newblock In \emph{Proceedings of the 2018 Conference on Empirical Methods in
  Natural Language Processing}, pages 22--32, Brussels, Belgium. Association
  for Computational Linguistics.

\bibitem[{Pradeep et~al.(2021{\natexlab{a}})Pradeep, Ma, Nogueira, and
  Lin}]{pradeep-etal-2021-scientific}
Ronak Pradeep, Xueguang Ma, Rodrigo Nogueira, and Jimmy Lin.
  2021{\natexlab{a}}.
\newblock \href {https://aclanthology.org/2021.louhi-1.11} {Scientific claim
  verification with {V}er{T}5erini}.
\newblock In \emph{Proceedings of the 12th International Workshop on Health
  Text Mining and Information Analysis}, pages 94--103, online. Association for
  Computational Linguistics.

\bibitem[{Pradeep et~al.(2021{\natexlab{b}})Pradeep, Ma, Nogueira, and
  Lin}]{pradeep2021vera}
Ronak Pradeep, Xueguang Ma, Rodrigo Nogueira, and Jimmy Lin.
  2021{\natexlab{b}}.
\newblock Vera: Prediction techniques for reducing harmful misinformation in
  consumer health search.
\newblock In \emph{Proceedings of the 44th International ACM SIGIR Conference
  on Research and Development in Information Retrieval}, pages 2066--2070.

\bibitem[{Rajpurkar et~al.(2016)Rajpurkar, Zhang, Lopyrev, and
  Liang}]{rajpurkar-etal-2016-squad}
Pranav Rajpurkar, Jian Zhang, Konstantin Lopyrev, and Percy Liang. 2016.
\newblock \href {https://doi.org/10.18653/v1/D16-1264} {{SQ}u{AD}: 100,000+
  questions for machine comprehension of text}.
\newblock In \emph{Proceedings of the 2016 Conference on Empirical Methods in
  Natural Language Processing}, pages 2383--2392, Austin, Texas. Association
  for Computational Linguistics.

\bibitem[{Robertson and Zaragoza(2009)}]{robertson2009probabilistic}
Stephen Robertson and Hugo Zaragoza. 2009.
\newblock The probabilistic relevance framework: {BM25} and beyond.
\newblock \emph{Foundations and Trends in Information Retrieval},
  3(4):333--389.

\bibitem[{Schlichtkrull et~al.(2021)Schlichtkrull, Karpukhin, Oguz, Lewis, Yih,
  and Riedel}]{schlichtkrull-etal-2021-joint}
Michael~Sejr Schlichtkrull, Vladimir Karpukhin, Barlas Oguz, Mike Lewis,
  Wen-tau Yih, and Sebastian Riedel. 2021.
\newblock \href {https://doi.org/10.18653/v1/2021.acl-long.529} {Joint
  verification and reranking for open fact checking over tables}.
\newblock In \emph{Proceedings of the 59th Annual Meeting of the Association
  for Computational Linguistics and the 11th International Joint Conference on
  Natural Language Processing (Volume 1: Long Papers)}, pages 6787--6799,
  Online. Association for Computational Linguistics.

\bibitem[{Schuster et~al.(2021)Schuster, Fisch, and
  Barzilay}]{schuster-etal-2021-get}
Tal Schuster, Adam Fisch, and Regina Barzilay. 2021.
\newblock \href {https://doi.org/10.18653/v1/2021.naacl-main.52} {Get your
  vitamin {C}! robust fact verification with contrastive evidence}.
\newblock In \emph{Proceedings of the 2021 Conference of the North American
  Chapter of the Association for Computational Linguistics: Human Language
  Technologies}, pages 624--643, Online. Association for Computational
  Linguistics.

\bibitem[{Seifert(2002)}]{seifert2002continued}
Colleen~M Seifert. 2002.
\newblock The continued influence of misinformation in memory: What makes a
  correction effective?
\newblock In \emph{Psychology of learning and motivation}, volume~41, pages
  265--292. Elsevier.

\bibitem[{Shah et~al.(2020)Shah, Schuster, and Barzilay}]{shah2020automatic}
Darsh Shah, Tal Schuster, and Regina Barzilay. 2020.
\newblock Automatic fact-guided sentence modification.
\newblock In \emph{Proceedings of the AAAI Conference on Artificial
  Intelligence}, volume~34, pages 8791--8798.

\bibitem[{Soleimani et~al.(2020)Soleimani, Monz, and
  Worring}]{soleimani2020bert}
Amir Soleimani, Christof Monz, and Marcel Worring. 2020.
\newblock Bert for evidence retrieval and claim verification.
\newblock In \emph{European Conference on Information Retrieval}, pages
  359--366. Springer.

\bibitem[{Srivastava et~al.(2014)Srivastava, Hinton, Krizhevsky, Sutskever, and
  Salakhutdinov}]{srivastava2014dropout}
Nitish Srivastava, Geoffrey Hinton, Alex Krizhevsky, Ilya Sutskever, and Ruslan
  Salakhutdinov. 2014.
\newblock Dropout: a simple way to prevent neural networks from overfitting.
\newblock \emph{The journal of machine learning research}, 15(1):1929--1958.

\bibitem[{Stammbach(2021)}]{stammbach-2021-evidence}
Dominik Stammbach. 2021.
\newblock \href {https://doi.org/10.18653/v1/2021.fever-1.2} {Evidence
  selection as a token-level prediction task}.
\newblock In \emph{Proceedings of the Fourth Workshop on Fact Extraction and
  VERification (FEVER)}, pages 14--20, Dominican Republic. Association for
  Computational Linguistics.

\bibitem[{Stammbach and Neumann(2019)}]{stammbach-neumann-2019-team}
Dominik Stammbach and Guenter Neumann. 2019.
\newblock \href {https://doi.org/10.18653/v1/D19-6616} {Team {DOMLIN}:
  Exploiting evidence enhancement for the {FEVER} shared task}.
\newblock In \emph{Proceedings of the Second Workshop on Fact Extraction and
  VERification (FEVER)}, pages 105--109, Hong Kong, China. Association for
  Computational Linguistics.

\bibitem[{Subramanian and Lee(2020)}]{subramanian-lee-2020-hierarchical}
Shyam Subramanian and Kyumin Lee. 2020.
\newblock \href {https://doi.org/10.18653/v1/2020.emnlp-main.627}
  {{H}ierarchical {E}vidence {S}et {M}odeling for automated fact extraction and
  verification}.
\newblock In \emph{Proceedings of the 2020 Conference on Empirical Methods in
  Natural Language Processing (EMNLP)}, pages 7798--7809, Online. Association
  for Computational Linguistics.

\bibitem[{Sundararajan et~al.(2017)Sundararajan, Taly, and
  Yan}]{sundararajan2017axiomatic}
Mukund Sundararajan, Ankur Taly, and Qiqi Yan. 2017.
\newblock Axiomatic attribution for deep networks.
\newblock In \emph{International conference on machine learning}, pages
  3319--3328. PMLR.

\bibitem[{Thorne et~al.(2021)Thorne, Glockner, Vallejo, Vlachos, and
  Gurevych}]{thorne2021evidence}
James Thorne, Max Glockner, Gisela Vallejo, Andreas Vlachos, and Iryna
  Gurevych. 2021.
\newblock \href {https://arxiv.org/abs/2104.00640} {Evidence-based verification
  for real world information needs}.
\newblock \emph{ArXiv preprint}, abs/2104.00640.

\bibitem[{Thorne and Vlachos(2021)}]{thorne-vlachos-2021-evidence}
James Thorne and Andreas Vlachos. 2021.
\newblock \href {https://doi.org/10.18653/v1/2021.acl-long.256} {Evidence-based
  factual error correction}.
\newblock In \emph{Proceedings of the 59th Annual Meeting of the Association
  for Computational Linguistics and the 11th International Joint Conference on
  Natural Language Processing (Volume 1: Long Papers)}, pages 3298--3309,
  Online. Association for Computational Linguistics.

\bibitem[{Thorne et~al.(2018)Thorne, Vlachos, Christodoulopoulos, and
  Mittal}]{thorne-etal-2018-fever}
James Thorne, Andreas Vlachos, Christos Christodoulopoulos, and Arpit Mittal.
  2018.
\newblock \href {https://doi.org/10.18653/v1/N18-1074} {{FEVER}: a large-scale
  dataset for fact extraction and {VER}ification}.
\newblock In \emph{Proceedings of the 2018 Conference of the North {A}merican
  Chapter of the Association for Computational Linguistics: Human Language
  Technologies, Volume 1 (Long Papers)}, pages 809--819, New Orleans,
  Louisiana. Association for Computational Linguistics.

\bibitem[{Uscinski and Butler(2013)}]{uscinski2013epistemology}
Joseph~E Uscinski and Ryden~W Butler. 2013.
\newblock The epistemology of fact checking.
\newblock \emph{Critical Review}, 25(2):162--180.

\bibitem[{Vaswani et~al.(2017)Vaswani, Shazeer, Parmar, Uszkoreit, Jones,
  Gomez, Kaiser, and Polosukhin}]{NIPS2017_transformers}
Ashish Vaswani, Noam Shazeer, Niki Parmar, Jakob Uszkoreit, Llion Jones,
  Aidan~N Gomez, \L~ukasz Kaiser, and Illia Polosukhin. 2017.
\newblock \href
  {https://proceedings.neurips.cc/paper/2017/file/3f5ee243547dee91fbd053c1c4a845aa-Paper.pdf}
  {Attention is all you need}.
\newblock In \emph{Advances in Neural Information Processing Systems},
  volume~30. Curran Associates, Inc.

\bibitem[{Williams et~al.(2018)Williams, Nangia, and Bowman}]{mnli}
Adina Williams, Nikita Nangia, and Samuel Bowman. 2018.
\newblock \href {http://aclweb.org/anthology/N18-1101} {A broad-coverage
  challenge corpus for sentence understanding through inference}.
\newblock In \emph{Proceedings of the 2018 Conference of the North American
  Chapter of the Association for Computational Linguistics: Human Language
  Technologies, Volume 1 (Long Papers)}, pages 1112--1122. Association for
  Computational Linguistics.

\bibitem[{Wolf et~al.(2019)Wolf, Debut, Sanh, Chaumond, Delangue, Moi, Cistac,
  Rault, Louf, Funtowicz et~al.}]{wolf2019huggingface}
Thomas Wolf, Lysandre Debut, Victor Sanh, Julien Chaumond, Clement Delangue,
  Anthony Moi, Pierric Cistac, Tim Rault, R{\'e}mi Louf, Morgan Funtowicz,
  et~al. 2019.
\newblock Huggingface's transformers: State-of-the-art natural language
  processing.
\newblock \emph{arXiv preprint arXiv:1910.03771}.

\bibitem[{Yin and Roth(2018)}]{yin-roth-2018-twowingos}
Wenpeng Yin and Dan Roth. 2018.
\newblock \href {https://doi.org/10.18653/v1/D18-1010} {{T}wo{W}ing{OS}: A
  two-wing optimization strategy for evidential claim verification}.
\newblock In \emph{Proceedings of the 2018 Conference on Empirical Methods in
  Natural Language Processing}, pages 105--114, Brussels, Belgium. Association
  for Computational Linguistics.

\bibitem[{Zhong et~al.(2020)Zhong, Xu, Tang, Xu, Duan, Zhou, Wang, and
  Yin}]{zhong-etal-2020-reasoning}
Wanjun Zhong, Jingjing Xu, Duyu Tang, Zenan Xu, Nan Duan, Ming Zhou, Jiahai
  Wang, and Jian Yin. 2020.
\newblock \href {https://doi.org/10.18653/v1/2020.acl-main.549} {Reasoning over
  semantic-level graph for fact checking}.
\newblock In \emph{Proceedings of the 58th Annual Meeting of the Association
  for Computational Linguistics}, pages 6170--6180, Online. Association for
  Computational Linguistics.

\bibitem[{Zhou et~al.(2019)Zhou, Han, Yang, Liu, Wang, Li, and
  Sun}]{zhou-etal-2019-gear}
Jie Zhou, Xu~Han, Cheng Yang, Zhiyuan Liu, Lifeng Wang, Changcheng Li, and
  Maosong Sun. 2019.
\newblock \href {https://doi.org/10.18653/v1/P19-1085} {{GEAR}: Graph-based
  evidence aggregating and reasoning for fact verification}.
\newblock In \emph{Proceedings of the 57th Annual Meeting of the Association
  for Computational Linguistics}, pages 892--901, Florence, Italy. Association
  for Computational Linguistics.

\bibitem[{Zubiaga et~al.(2016)Zubiaga, Liakata, Procter, Hoi, and
  Tolmie}]{zubiaga2016analysing}
Arkaitz Zubiaga, Maria Liakata, Rob Procter, Geraldine Wong~Sak Hoi, and Peter
  Tolmie. 2016.
\newblock Analysing how people orient to and spread rumours in social media by
  looking at conversational threads.
\newblock \emph{PloS one}, 11(3):e0150989.

\end{thebibliography}

\appendix
\clearpage
\section{Performance on FAVIQ-A}
\label{app:FAVIQ}
\begin{table}[t]
    \centering
    \scalebox{0.8}{\begin{tabular}{lrrrr}
\hline
                          & \multicolumn{1}{c}{\textbf{Test}} & \multicolumn{1}{c}{\textbf{Dev}} & \multicolumn{1}{c}{$\boldsymbol{\Delta}$} & \multicolumn{1}{c}{$\theta$}\\ \hline
BART\small{$_{LARGE}$} \small{\cite{park-etal-2022-faviq} }        & 64.9                              & 66.9                            & 2.0               &           374M       \\ % roughly 10% more than BERT

FiD \small{\cite{asai-etal-2022-evidentiality} }        & 65.7                              & 69.6                             & 3.9            &336M                    \\
CD\small{$_{RoBERTa}$}         & 58.6                              & 69.8                             & 11.2                   & 127M           \\
CD\small{$_{RoBERTaL}$}    & 66.9                              & 73.3                             & 6.4                       &360M         \\
CD       & 69.8                              & 76.3                             & 6.5                               &187M \\
%CD \textbackslash{}wo RC & \multicolumn{1}{l}{}              & 73.3                             & \multicolumn{1}{l}{}               \\
CD\small{$_{LARGE}$}   & 72.0                                & 79.7                             & 7.7       & 439M \\ \hline                        
\end{tabular}
}%
    \caption{Performance on FAVIQ-A.}
    \label{fig:FAVIQ}
\end{table}
To asses more realistic performance of our system, we study its performance on FAVIQ-A \cite{park-etal-2022-faviq}. We use the silver passage supervision from \citet{asai-etal-2022-evidentiality}, and feed the model with top-20 passages retrieved via DPR system \cite{karpukhin-etal-2020-dense}. We keep all the hyperparameters same as for FEVER, and use dev set only for early-stopping. We compare to evidentiality-guided generator (EGG), a t5-based Fusion-in-Decoder (FiD) \cite{izacard-grave-2021-leveraging} with two decoders from \citet{asai-etal-2022-evidentiality}. We use hyperparameters from FEVER. The results are shown in Table \ref{fig:FAVIQ}. Our DebertaV3-based Claim-Dissector reaches state-of-the-art results on the dataset. The domain mismatch (measured by difference $\boldsymbol{\Delta}$) between development and test set is likely caused by the domain shift of NaturalQuestions test set, from which FAVIQ's test set was created (see Appendix B in \citet{min-etal-2020-ambigqa}). However, despite our best efforts, we have not uncovered the cause of massive degradation between dev and test set for \texttt{roberta-base} based Claim-Dissector (the standard deviation on test set was only $\pm0.4$ accuracy points).

\section{Performance on RealFC}
\label{app:REALFC}
\begin{table*}[t]
    \centering
    \scalebox{0.8}{\begin{tabular}{llrrrrr}
\toprule
\textbf{Dataset}                &   \textbf{Model}                                                       & \textbf{EviF1} & \textbf{VA} & \textbf{VF1} & \textbf{CondAcc} & \textbf{CondF1} \\ \midrule
\multirow{6}{*}{Full} & 3-way+$\mathbb{C}$ \small{\cite{thorne2021evidence} }    & 46.8                      & 75.5                     & 65.5                    & 64.2                        & 51.8                       \\
                      & Any-Best+$\mathbb{C}$ \small{\cite{thorne2021evidence} } & 48.6                      & 75.5                     & 65.5                    & 64.2                        & 52.2                       \\
                      & CD\small{$_{RoBERTa}$}\textbackslash{}W                  & 48.7                      & 76.7                     & 66.6                    & 64.8                        & 52.4                       \\
                      & CD\small{$_{RoBERTa}$}                                   & 53.0                      & 76.3                     & 68.6                    & 65.0                        & 55.3                       \\
                      & CD                                                       & 54.7                      & 79.1                     & 72.2                    & 67.6                        & 58.5                       \\
                      & CD\small{$_{LARGE}$}                                     & 56.3                      & 80.8                     & 73.7                    & 69.6                        & 60.7                       \\ \midrule
\multirow{2}{*}{Bipolar} & 3-way+$\mathbb{C}$ \small{\cite{thorne2021evidence} } & 48.1$\pm3.3$              & 43.0$\pm1.1$             & 40.3$\pm1.2$           & 23.0$\pm1.4$                & 28.7$\pm1.4$                \\
                      & CD\small{$_{RoBERTa}$}                                   & 56.4$\pm0.7$              & 51.9$\pm1.8$             & 47.7$\pm2.9$            & 32.4$\pm0.6$                & 38.7$\pm1.2$               \\\bottomrule
\end{tabular}}%
    \caption{Performance on RealFC test set.}
    \label{tab:cd_realfc}
\end{table*}

To analyze, whether our token-level interpretability approach transfers beyond FEVER dataset, and how our model performs in bipolar evidence setting (when claim is both supported by some evidence, but refuted by a different one), we study our model on RealFC \cite{thorne2021evidence}\footnote{Unlike FaVIQ, this dataset also contains sentence-level annotation.}.

Similarly to {FS} metric on FEVER, RealFC dataset is validated through conditional scoring. It validates both; the performance of reranking and veracity prediction. Specifically, it computes average accuracy/F1 score across samples, while setting per-sample hit/{F1} to 0, if the model predicted the wrong veracity. Due to a rather complicated exact definition, official conditional scores are documented in Appendix~\ref{app:conditiona-scores}.

The results on RealFC are shown in Table~\ref{tab:cd_realfc}. For relevant evidence classification, binary {F1} computed from the concatenation of all relevant/irrelevant decisions for all sentences is reported. Here supporting or refuting evidence counts as relevant, neutral as irrelevant. For verification, accuracy/macro {F1} (denoted {VAcc}, {VF1}) are reported. Lastly, we also report conditional scores {CondAcc} and {CondF1}, which are aggregated from per-sample binary {F1} relevance, set to 0 if the target veracity was not predicted correctly\footnote{Due to rather complicated exact definition, official conditional scores are documented in Appendix~\ref{app:conditiona-scores}.}. We compare CD with a pipelined baseline composed from a 3-class relevance classification (assuming either supporting or refuting evidence is relevant) followed by a veracity classifier (first row). Both components are based on \texttt{roberta-base} \cite{liu2019roberta}. We also report best number for the corresponding column across all (RoBERTa-based) baselines from \citet{thorne2021evidence} (second row). We find that in a similar setup, CD improves only marginally over baseline. The early stopping of baselines and the third row is performed on {VAcc}. Row 4 and further report on results early stopped on CondF1, as we found CondF1 to correlate with the majority of the metrics. To counter data imbalance, from row 4, we use veracity class weighting. Using identity-weighting (each weight is 1), we observe accuracy to be maximal, whereas using the inverse-class-prior weighting exactly as in the previous chapter we found {F1} to perform the best. However, for maximizing conditional scores, we found the weighting corresponding to average of identity-weighting and inverse-class-prior to work the best, and so we report these further.

We find that CD-based systems retain its interpretability\footnote{Qualitative samples with token-level rationales on bipolar subset of RealFC are available at \url{https://shorturl.at/hGU67}.} with class weighting sets a new state-of-the-art on the dataset. Additionally, we uncover a large performance boost on the subset of the dataset with annotated {bipolar evidence} (last two rows).
\section{Performance on HoVer}
\label{app:hover_res}
\begin{table}
    \centering
    \scalebox{1.}{\begin{tabular}{cllr}
 & Hops                     & \textbf{Accuracy} & \multicolumn{1}{l}{\textbf{EM}} \\ \hline 
\multirow{5}{*}{\rotatebox[origin=c]{90}{Baleen}}
& \multicolumn{1}{l|}{2}    & -                 & 47.3                            \\
& \multicolumn{1}{l|}{3}    & -                 & 37.7                            \\
& \multicolumn{1}{l|}{4}    & -                 & 33.3                            \\
& \multicolumn{1}{l|}{All}  & 84.5              & 39.2                            \\

\hline
\multirow{5}{*}{\rotatebox[origin=c]{90}{CD}}
& \multicolumn{1}{l|}{2}    & 81.3              & 48.0                            \\
& \multicolumn{1}{l|}{3}    & 80.1              & 16.9                            \\
& \multicolumn{1}{l|}{4}    & 78.1              &  7.7                            \\
& \multicolumn{1}{l|}{All} & 79.9              & 23.3                            \\

\end{tabular}}%
    \caption{Results on HoVer dataset (dev split).}
    \label{fig:hover_res}
\end{table}
To study how well our model can deal with claims, which require multihop information, we trained our system on HoVer \cite{jiang-etal-2020-hover}. In particular, we follow the recipe for Baleen \cite{khattab2021baleen} and retrieve 4$\times$25 top articles using official quantized Baleen implementation\footnote{\url{https://github.com/stanford-futuredata/Baleen}} (which achieves about 2\,\% lower retrieval@100 on supported samples than reported in paper). We split 5 starting documents from each iteration into blocks, padding input with further documents from first retrieval iteration when necessary. We keep input size at $K_1=35$, and we do not use hyperlink expansion. We compute the probability of not-supported class by summing NEI and REFUTE classes. Furthermore, we assume simplified conditions, we infuse inputs with oracle information when necessary (achieving RaI ~100\,\%) and predict as many evidences, as there was annotated. We refer reader to \citet{khattab2021baleen} for further information about setup and evaluation metrics.

Nevertheless, our system lags behind Baleen on 3 and 4 hop examples, as shown in Table \ref{fig:hover_res}. We hyphothesize that, similarly to Baleen, autoregressive process is necessary to match its performance. We leave the question of interpretable multi-hop fact-checking with Claim-Dissector open for our future work.
% \vfill
% \pagebreak
% \begin{figure}[ht]
%     \centering
%     \includegraphics[width=\linewidth]{figures/cd_example.png}
%     \caption{Example of interpretable refuting evidence from Claim-Dissector for claim ``\textit{American Sniper (book) is about a sniper with 170 officially confirmed kills.}''.}
%     \label{fig:intexample}
% \end{figure}
\section{Detection of examples with bipolar evidence.}
\label{app:bipolarevi}
We manually analyzed whether we can take advantage of model's ability to distinguish between evidence, which is relevant because it supports the claim, and the evidence which is relevant because it refutes the claim. To do so, we try to automatically detect examples from the validation set, which contain both, supporting and refuting evidence (which we refer to as bipolar evidence). We note that there were no examples with explicitly annotated bipolar evidence in the training data. 

% Formally we select every example for which ~$\exists a,b,x,y:  \operatorname{P}^{a,b}({y}=S)>0.9 \land \operatorname{P}^{x,y}({y}=R)>0.9$.
% Hidden for review
We select all examples where model predicted at least~$0.9$ probability for any supporting and any refuting evidence. We found that out of~$72$ such examples,~$66\,\%(48)$ we judged as indeed having the bipolar evidence\footnote{Annotations are available at \url{shorturl.at/qrtIP}.}.
We observed that about half ($25$/$48$) of these examples had bipolar evidence because of the entity ambiguity caused by open-domain setting. E.g., claim ``\textit{Bones is a movie}'' was supported by sentence article ``\textit{Bones (2001 film)}'' but also refuted by sentence from article ``\textit{Bones (TV series)}'' and ``\textit{Bone}'' (a rigid organ).

\section{Conditional Scoring on RealFC Dataset}
\label{app:conditiona-scores}
First, binary F1 is computed for each $i$-th example from the dataset---section with several sentences, and every sentence is predicted to be relevant (i.e., supporting or refuting) or irrelevant (i.e., neutral)---obtaining F1 score $s(i)$. Define $\mathbb{I}[m(i)=l_i]$ as an indicator function yielding 1 if model $m_v$ predicted the veracity correct class $l_i$ for an $i$-th example, and 0 otherwise. With dataset of size $N$, the {CondAcc} is defined as an \emph{average F1 score for samples with correct veracity}
\begin{equation}
    \operatorname{{CondAcc}}=\frac{1}{N}\sum_{i=1}^N s(i)\mathbb{I}[m_v(i)=l_i].
\end{equation}
Next, assume that $TP_s$, $TP_r$, and $TP_n$  and similarly $\operatorname{FP}_x$ and $\operatorname{FN}_x$ are TPs, FPs, and FNs computed for each veracity class---support veracity ($s$), refute veracity ($r$), or neutral veracity ($n$), and $x\in\{s,r,n\}$. Further, assume that $wTP_s$, $wTP_r$, $wTP_n$ are computed as $wTP_x = \sum_{i=1}^N s(i)\mathbb{I}[m_v(i)=l_i \wedge x=l_i]$, where $x\in\{s,r,n\}$. Firstly, weighted F1 $wF1_x$ is computed for each class separately from precision $p_x$ and recall $r_x$ as shown in Equation~\ref{eq:weightedF1}. The {CondF1} score is obtained by averaging $wF1_x$ across all classes $x\in\{s,r,n\}$.
\begin{equation}
\label{eq:weightedF1}
\begin{split}
	  p_x= \frac{wTP_x}{TP_x + FP_x}\qquad
	r_x= \frac{wTP_x}{TP_x + FN_x}\\
	wF1_x= 2\frac{p_x r_x}{p_x+r_x}
\end{split}
\end{equation}
\section{Sensitivity to \texorpdfstring{$\lambda_2$}{lambda2} Weight}
\label{app:l2_sensitivity}
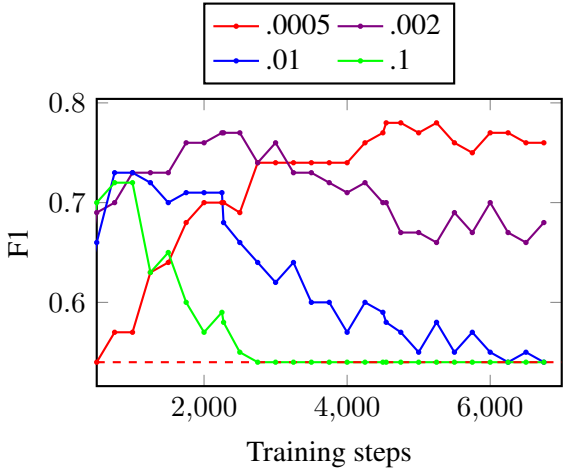
\begin{figure}
    \centering
    \begin{tikzpicture}
    \begin{axis}[
            %smooth,
            thick,
            legend columns=2,
            legend style={
                at={(0.5,1.05)}, 
                legend cell align={left},  
                anchor=south
            },
            xmin=500, xmax=7000,
            xlabel={Training steps},
            ylabel=F1,
            width=1.0\columnwidth,
            height=0.7\columnwidth,
            % xtick=data,
            % xmajorgrids,
            % ymajorgrids,
        ]%Steps,FS1,FS2,FS3,Average FS,F11,F12,F13,Average F1
        \addplot+[solid, mark=*, red,mark options={scale=0.3}]               table [x=STEPS, y=.0005, col sep=comma] {data/L2_weight_penalty.csv};
        \addlegendentry{.0005}
        \addplot+[solid, mark=*, violet,mark options={scale=0.3}]               table [x=STEPS, y=.002, col sep=comma] {data/L2_weight_penalty.csv};
        \addlegendentry{.002}
        \addplot+[solid, mark=*, blue,mark options={scale=0.3}]               table [x=STEPS, y=.01, col sep=comma] {data/L2_weight_penalty.csv};
        \addlegendentry{.01}
        \addplot+[solid, mark=*, green,mark options={scale=0.3}]               table [x=STEPS, y=.1, col sep=comma] {data/L2_weight_penalty.csv};
        \addlegendentry{.1}       
        \addplot[dashed,thick, samples=50, smooth,domain=0:6,red] coordinates {(500,0.54)(7000,0.54)};

    \end{axis}%
\end{tikzpicture}%
    \caption{Sensitivity to $\lambda_2$ weight selection for Deberta-V3-base model. Red dashed horizontal line marks the F1 performance when selecting all tokens (Select All Tokens) from Table \ref{tab:slr_to_tlr}.}
    \label{fig:l2_sensitivity}
\end{figure}
In Figure \ref{fig:l2_sensitivity}, we analyze the sensitivity of $\lambda_2$ parameter on F1 performance on TLR-FEVER during training. Choosing the large weight may lead to instabilities and vanishing of interpretable rationales, choosing smaller weight delays the peak performance in terms of F1. Note that for DeBERTaV3-large we chose $\lambda_2=.0005$, as the one we used for base version ($\lambda_2=.002$) leaded to such vanishing.
\section{Structure of Single-layer Perceptron}
\label{sec:app_SLP}
Given a vector~$\boldsymbol{x}$, the structure of single-layer perceptron from equation~\ref{eq:scorecomp} is the following:
\begin{equation}
    \operatorname{SLP}(\boldsymbol{x}) = \operatorname{GELU}(\operatorname{dp}(\boldsymbol{W'}\operatorname{lnorm}(\boldsymbol{x}))).
\end{equation}
The operator~$\operatorname{dp}$ denotes the dropout \cite{srivastava2014dropout} used in training,~$\boldsymbol{W}'$ is a trainable matrix,~$\operatorname{GELU}$ is the Gaussian Error Linear Unit \cite{hendrycks2016gaussian} and~$\operatorname{lnorm}$ is the layer normalization \cite{ba2016layer}.

\section{Details on Analysis of Prediction Scores}
\label{app:psuseful}
We define relevance score (RS) as $\operatorname{P}^{i,j}({y}=l)$ where $l\in\{S,R\}$ is the ground truth label. For A/B testing, we shuffle (a) and (b) cases for annotators, so they cannot determine which sentence comes from which source. An example of 1 annotation is available at \url{https://shorturl.at/hiCLX}. Since we found many annotators hesitated with no preference option (c) when computing Krippendroff's $\alpha$, we assume two classes, and empty preference when the annotator does not have a preference (case (c)) (we do not consider it a separate nominal category). Krippendorff's $\alpha$=0.65 achieved moderate agreement.

To compute average Kendall $\tau$, we select sentences with RS>0.7 for each example; this creates relevance ranking. Then we reorder selected sentences according to PS, obtaining prediction ranking. Kendall $\tau$  is computed for every sample, and the resulting statistic is averaged across all samples in the FEVER validation set. 

\section{Experiments with Different Model Parametrizations}
\label{sec:app_diffparam}
Apart from parametrizations provided in the main paper, we experimented with several different parametrizations described below. We keep the training details the same as for our baseline (Section~\ref{section:baseline}). Starting off with a baseline system formulation, we will consider replacing~$L_{b0}$ with different objective functions.
\begin{equation}
    \mathcal{L}_{b2} = \frac{1}{|\mathbb{A}|}\sum_{{s}_{i,j}, {y} \in \mathbb{A}} \log \operatorname{P}({s}_{i,j}, {y}) 
\end{equation}

With~$L_{b2}$, the annotation set~$\mathbb{A}$ contains both relevant and irrelevant annotations. We found in practice this does not work - recall@5 during training stays at 0. This makes sense since if annotation exists, the final class is likely support or refute. Drifting the probability mass towards NEI for irrelevant annotations is adversarial w.r.t. total veracity probability.

\begin{equation}
    \mathcal{L}_{b3} = \log\sum_{{s}_{i,j}, {y} \in \mathbb{A}_p}  \operatorname{P}({s}_{i,j}, {y}) 
\end{equation}
Instead of maximizing the multinomial probability,~$L_{b3}$ objective marginalizes over relevant annotations.

\begin{equation}
    \mathcal{L}_{b4} = \log\sum_{{s}_{i,j}\in \mathbb{A}_p}\sum_{{y}}  \operatorname{P}({s}_{i,j}, {y}) 
\end{equation}
Additionally to~$L_{b3}$,~$L_{b4}$ also marginalizes out the class label~${y}$.

\begin{table}[t]
    \centering
    \scalebox{0.75}{% \begin{tabular}{lccc|ccc}
%     & \multicolumn{3}{c}{\textbf{FEVER}}   & \multicolumn{3}{c}{\textbf{FEVER$_{MH}$}}      \\ 
%     \Xhline{4\arrayrulewidth}
% \textbf{Model}                     & \textbf{FS} & \textbf{A} & \textbf{R@5} & \textbf{FS} & \textbf{A} & \textbf{R@5} \\ \hline
% CD                                 & 76.2       & 79.5      & 91.5          & 26.3        & 68.9      & 35.5         \\
% Baseline                           & 75.2       & 79.8      & 90.0          & 26.1        & 74.2      & 33.9         \\
% $L_{b3}$                           & 76.0       & 79.0      & 91.2          & 20.2        & 71.8      & 26.3        \\
% $L_{b4}$                           & 75.7       & 79.7      & 90.4          & 23.4        & 72.3      & 31.4        \\ \hline
% \end{tabular}
\begin{tabular}{lccc}
    & \multicolumn{3}{c}{\textbf{FEVER}}   \\ 
    \Xhline{4\arrayrulewidth}
\textbf{Model}                     & \textbf{FS} & \textbf{A} & \textbf{R@5}  \\ \hline
CD                                 & 76.2       & 79.5      & 91.5         \\
Baseline                           & 75.2       & 79.8      & 90.0         \\
$L_{b3}$                           & 76.0       & 79.0      & 91.2         \\
$L_{b4}$                           & 75.7       & 79.7      & 90.4         \\ \hline
\end{tabular}}%
    \caption{Different model parametrizations.}
    \label{fig:ablation_losses}
\end{table}
The results in Table~\ref{fig:ablation_losses} reveal only minor differences. Comparing~$L_{b3}$ and~$L_{b4}$, marginalizing out label improves the accuracy, but damages the recall. Baseline parametrization achieves best accuracy but the worst recall. Claim-Dissector seems to work the best in terms of FS, but the difference to~$L_{b3}$ is negligible, if any.

\section{Statistical Testing on F1 Measure}
\label{app:stat_testing_f1}
To compare CD with masker in F1, we follow \citet{goutte2005probabilistic}, sum TPs, FPs, FNs across the dataset, estimate recall (R) and precision (P) posteriors, and sample F1 distributions. To obtain sample of average F1 from multiple checkpoints, we estimate the P and R posteriors for each checkpoint separately, sample F1 for each checkpoint and then average these. We estimate $p \approx \operatorname{P}($F1$_a>$F1$_b)$ via Monte-Carlo, and consider significance level at $p>0.95$. 

\section{The Continued Influence Effect: Retractions
Fail to Eliminate the Influence of
Misinformation}
\label{app:lewandovsky}
\citet{lewandowsky2012misinformation} summarizes research paradigm, which focuses on credible retractions in neutral scenarios, in which people have no reason to believe one version of the event over another. In this paradigm, people are presented with a factious report about an event unfolding over time. The report contain a target piece of information (i.e. a claim). For some readers, the claim is retracted, whereas for readers in a control condition, no correction occurs. Then the readers are presented with a questionnare to assess their understanding of the event and the number of clear and uncontroverted references to the claim's veracity.

In particular, a stimulus narrative commonly used within this paradigm involves a warehouse fire, that is initially thought to have been caused by gas cylinders and oil paints there were negligently stored in a closet. A proportion of participants is then presented with a retraction such as "the closet was actually empty". A comprehension test follows, and number of references to the gas and paint in response to indirection inference questions about the event (e.g., "What caused the black smoke?") is counted. 

Research using this paradigm has consistently found that retractions rarely, if ever, had the intended effect of eliminating reliance on misinformation, even when people remember the retraction, when later asked. 
\citet{seifert2002continued} have examined whether clarifying the correction might reduce the continued influence effect. The correction in their studies was strengthened to include the phrase "paint and gas were never on the premises". Results showed that this enhanced negation of the presence of flammable materials backfired, making people even more likely to rely on the misinformation in their responses. Some other additions to the correction were found to mitigate to a degree, but not eliminate, the continued influence effect. For instance, when participants were given a rationale about how misinformation originated, such as "a truckers' strike prevented the expected delivery of the items", they were less likely to make references to it. Even so, the influence of the misinformation could still be detected. The conclusion drawn from studies on this phenomenon show that it is extremely difficult to return the beliefs to people who have been exposed to misinformation to a baseline similar to those of people who have never been exposed to it. We recommend reading \citet{lewandowsky2012misinformation} for broader overview of \textit{the misinformation and its correction}.

\section{Masker}
\label{app:masker}
\begin{description}[style=unboxed,leftmargin=0em,listparindent=\parindent]
    \setlength\parskip{0em}
\item[Model Description.]
Our masker follows same \mbox{DeBERTaV3} architecture as Claim-Dissector, except that the multiheaded layer from the equation \eqref{eq:scorecomp} is omitted. It receives~$K_1$ blocks at its input, encoded the very same way as for the Claim-Dissector. Instead of computing matrix~$\boldsymbol{M}$--- which contains three logits per evidence token, the masker predicts two logits~$[l^i_0,l^i_1]$ --- corresponding to keep/mask probabilities~$[p^i_0,p^i_1]$ for~$i$-th token in evidence of every block. The mask~$[m^i_0,m^i_1]$ is then sampled for every token from concrete distribution via Gumbel-softmax \cite{jang2017categorical}. During training,~$i$-th token embedding~$e_i$ at the Claim-Dissector's input~$e'_i$ is replaced with a linear combination of itself and a learned mask-embedding~$e_m \in \mathbb{R}^d$, tuned with the masker.
\begin{equation}
    e'_i = m^i_0 e_i + m^i_1e_m
\end{equation}
The masker is trained to maximize the Claim-Dissector's log-likehood of NEI class, while forcing the mask to be sparse via L1 regularization. Per-sample objective to maximize with sparsity strength hyperparameter~$\lambda_S$ is given as
\begin{equation}
    \mathcal{L} = \log\operatorname{P}({y} = NEI) -   \frac{\lambda_S}{L_e}\sum_i |m^i_0| .
\end{equation}

\item[Training Details.]
We keep most hyperparameters the same as for Claim-Dissector. The only difference is learning rate~$2e-6$, adaptive scheduling on Gumbel-softmax temperature~$\tau$ and training model/masker on different dataset split. Training starts with temperature~$\tau=1$ and after initial~$100$ steps, it is linearly decreasing towards~$\tau=0.1$ at step~$700$. Then we switch to hard Gumbel-softmax --- sampling 1-hot vectors in forward pass, while computing gradients as we would use a soft sample with ~$\tau=0.1$ at backward pass. We randomly split training set and we train model on 75\,\% of data, and masker on remaining 25\,\% of data. 

\end{description}
\section{Retrieval Performance}
\label{app:retrieval_perofmrnace}

% Retrieval Performance
\begin{table}[t]
    \centering
    \resizebox{\linewidth}{!}{\begin{tabular}{lcccr}
\hline
\textbf{K$_1$+K$_2$}  & \textbf{FEVER} & \textbf{FEVER}$_{MH}$ & \textbf{FEVER}$_{{MH}_{ART}}$ & \textbf{\#SaI} \\ \hline
35+0  & 94.2   & 52.0     & 45.8        & 239.9 \\
100+0 & 95.1   & 58.5     & 53.1        & 649.4 \\ 
35+10          & 95.2            & 61.9              & 57.0                 & 269.6          \\
35+20          & 95.9            & 69.0              & 65.2                 & 309.0          \\
35+30          & 96.7            & 77.5              & 74.7                 & 388.6          \\
35+35          & 97.5            & 84.1              & 82.3                 & 506.7          \\
35+50          & 97.7            & 86.5              & 85.0                 & 624.3          \\
35+100         & 98.4            & 93.0              & 92.4                 & 1008.8         \\
100+100        & 98.6            & 93.4              & 92.7                 & 1431.0         \\ \hline
\end{tabular}}%
    \caption{Retrieval performance in RaI on FEVER dev set and its subsets.}
    \label{fig:retrieval} 
\end{table}
We evaluate the retrieval method from \citet{jiang-etal-2021-exploring-listwise} and the proposed hyperlink expansion method in Table~\ref{fig:retrieval}. We use two metrics:
\begin{description}[style=unboxed,leftmargin=0em,listparindent=\parindent]
    \setlength\parskip{0em}
    \item [Recall@Input (RaI).] We evaluate retrieval w.r.t. recall at model's input while considering different amount of K$_1$+K$_2$ blocks at the input, i.e. the score hit counts iff any annotated evidence group was matched in K$_1$+K$_2$ input blocks.
    \item [Number of Sentences@Input (\#SaI)] denotes average number of sentences at model's input under corresponding~$K_1+K_2$ setting.
\end{description}
We focus on analyzing the effect of hyperlink expansion, varying~$K_2$, while keeping~$K_1 = 35$ in most experiments, which is setting similar to previous work --- \citet{jiang-etal-2021-exploring-listwise} considers reranking top-200 sentences. We observe that setting~$K_1 + K_2 = 35+10$ already outperforms retrieval without hyperlink expansion and~$K_1=100$ blocks. Such observation is thus consistent with previous work which used hyperlink signal \cite{hanselowski2018ukp,stammbach-neumann-2019-team}.

\begin{table}[ht]
    \centering
    \resizebox{\linewidth}{!}{\begin{tabular}{lllll}
\hline
\textbf{K}$_1$  & \textbf{Recall} & \textbf{Recall}$_{MH}$ & \textbf{Recall}$_{MH_{ART}}$ & \textbf{\#SaI} \\ \hline
10  & 90.4   & 40.1     & 33.0        & 68.8  \\
20  & 93.4   & 48.0     & 41.5        & 132.9 \\
30  & 94.1   & 51.3     & 45.0        & 196.8 \\
35  & 94.2   & 52.0     & 45.8        & 239.9 \\
50  & 94.5   & 54.3     & 48.4        & 325.4 \\
100 & 95.1   & 58.5     & 53.1        & 649.4 \\ \hline
\end{tabular}}%
    \caption{Retrieval performance on FEVER dev set.}
    \label{fig:retrieval_performance}
\end{table}
An in-depth evaluation of retrieval method adopted from \citet{jiang-etal-2021-exploring-listwise} is available in Table~\ref{fig:retrieval_performance}.
% \begin{table}[t]
%     \centering
%     \resizebox{\linewidth}{!}{\input{tables/he_performance}}%
%     \caption{Hyperlink expansion performance.}
%     \label{fig:ablations}
% \end{table}
\section{Experimental Details}
\label{app:experimental_details}
We base our implementation of pre-trained language representation models on Huggingface \cite{wolf2019huggingface}. Unless said otherwise, we employ \mbox{DeBERTaV3} \cite{he2021debertav3} as LRM. In all experiments, we firstly pre-train model on MNLI \cite{mnli}. While we observed no significant improvement when using a MNLI-pretrained checkpoint, we found that without MNLI pretraining, our framework sometimes converges to poor performance. We train model on FEVER with minibatch size~$64$, learning rate~$5e-6$, maximum block-length~$L_x=500$. We schedule linear warmup of learning rate for first~$100$ steps and then keep constant learning rate. We use Adam with decoupled weight decay \cite{loshchilov2017decoupled} and clip gradient vectors to a maximal norm of ~$1$ \cite{pascanu2013difficulty}. In all experiments, the model is trained and evaluated in mixed-precision. We keep~$\lambda_R = \lambda_2 =1$. We use 8x Nvidia A100 40GB GPUs for training. We validate our model every~$500$ steps and select best checkpoint according to FEVER-Score (see Subsection~\ref{ss:evaluation}). We have not used any principled way to tune the hyperparameters.

To train model with SSE, we decrease the strength of block-level supervised~$\mathcal{L}_R$ objective to~$\lambda_R=0.7$. We switch between vanilla objective and SSE objective randomly on per-sample basis. Training starts with replace probability~$p_{sse}=0$. for first~$1,000$ steps. The probability is then linearly increased up to~$p_{sse}=0.95$ on step~$3,000$, after which it is left constant.

All results except for Table \ref{fig:ablations} and Table \ref{tab:slr_to_tlr} were early-stopped based on the best FS. For Table \ref{fig:ablations}, we report best result for each metric early-stopped independently, to be comparable with ablations where FS was not available. For Table \ref{tab:slr_to_tlr}, we record best F1 during training.

% \section{Example of Predicted Rationales}
% \label{app:rationale_example}

\section{Logit Proof}
\label{app:logit_proof}
The link between equation \ref{eq:word_label_probability} and equation \ref{eq:logit_wordlabelprob} can be easily proved as follows. Applying logarithm to equation \ref{eq:word_label_probability} we get
\begin{equation}
        \log\operatorname{P}^{i,j}(w, y)={{\boldsymbol{M}^{i,j}_{w,y}}} - {\log\sum_{w'}\sum_{y'}\operatorname{exp}{\boldsymbol{M}^{i,j}_{w',y'}}}.
\end{equation}

Expressing $\boldsymbol{M}^{i,j}_{w,y}$, substituting $C^{i,j}=\sum_{w'}\sum_{y'}\operatorname{exp}{\boldsymbol{M}^{i,j}_{w',y'}}$, and merging the logarithms, we arrive to equation \ref{eq:logit_wordlabelprob}. We recommend \citet{bishop:2006:PRML}, chapter 4.2 for further information.

\section{The Necessity of Baseline Modifications}
\label{app:baselinemods}
The reason for the modification is that (i) the original model \cite{schlichtkrull-etal-2021-joint} (without $L_{b1}$) could not benefit from NEI annotations present on FEVER, resulting in unfair comparison with our models and previous work, as TabFact does not contain such annotations (ii) the original model is not able to distinguish the attribution from the repeated relevant evidence, because equations (6)/(9) in their work just sum the probabilities of relevant items supervised independently --- they do not use the supervision of overall veracity for the claim.
This is problematic especially in FEVER setting comparable to ours, where the relevance of hundreds of sentences is considered (many of them possibly relevant) as compared to TabFact where only top-5 retrieved tables were considered, and often only single is relevant.

\section{Token-level Annotation Guidelines}
\label{app:annotation_guidelines}
\textbf{Annotation Guidelines}\\
Welcome to the ``Pilot annotation phase'' and thank you for your help!\\
\textbf{How to start annotate}\\
If you haven't done so, simply click on "Start Annotation" button, and the annotation will start.\\
\textbf{Annotation process \& Guidelines}\\
\begin{itemize}
\item In pilot annotation, we are interested in annotator's disagreement on the task. So whatever disambiguity you will face, do not contact the organizers but judge it yourself.
\item Your task is to annotate 100 samples. In each case, you will be presented with list of sentences divided by | character. The sentences do not need to (and often do not) directly follow each other in text.
 Be sure that in each case you:
\item read the claim (lower-right corner)
\item read metadata - to understand the context, you also have access to other metadata (lower-right corner), such as
 \begin{itemize}
\item titles - Wikipedia article names for every sentence you are presented with, split with character |,
\item claim\_label - Ground-truth judgment of the claim's veracity.
\end{itemize}
\item \textbf{highlight minimal part of text you find important for supporting/refuting the claim. There should be such part in every sentence (unless annotation error happened). PLEASE DO NOT ANNOTATE ONLY WHAT IS IMPORTANT IN THE FIRST SENTENCE.}
\item Use \textbf{"RELEVANT"} annotation button highlight the selected text spans.
\item In some cases, you can find \textbf{errors in the ground-truth judgment}, in other words, either document is marked as supported and it should be refuted according to your judgment or vice-versa. If you notice so, please annotate any part of the document with \textbf{CLAIM\_ERROR} annotation.
\item In case you would like to comment on some example, use comment button (message icon). If the comment is example specific, please provide specific example's id (available in-between metadata).
 \end{itemize}
\textbf{FAQ}\\
\begin{itemize}
\item \textit{The example does not contain enough information to decide whether it should be supported or refuted. Should I label it as a CLAIM\_ERROR?}\\
 No. In such case, please annotate parts of the input, which are at least partially supporting or refuting the claim. Please add comment to such examples.
 If there are no such input parts, only then report the example as CLAIM\_ERROR.\\
 \end{itemize}
\textbf{That is it. Good luck!}

\end{document}